\documentclass{article}

\usepackage{PRIMEarxiv}

\usepackage[utf8]{inputenc} 
\usepackage[T1]{fontenc}    
\usepackage{hyperref}       
\usepackage{url}            
\usepackage{booktabs}       
\usepackage{amsfonts}       
\usepackage{nicefrac}       
\usepackage{microtype}      
\usepackage{lipsum}
\usepackage{fancyhdr}       
\usepackage{graphicx}       

\usepackage{amsmath}
\usepackage[capitalize]{cleveref}
\crefname{section}{Sec.}{Secs.}
\Crefname{section}{Section}{Sections}
\Crefname{table}{Table}{Tables}
\crefname{table}{Tab.}{Tabs.}
\usepackage{array}
\usepackage{makecell}
\usepackage{lipsum}  
\usepackage{multirow}
\usepackage{xcolor}
\usepackage{booktabs}  
\usepackage{makecell}
\usepackage{arydshln}
\usepackage{caption}
\usepackage{subcaption}
\usepackage{hyperref}
\usepackage{authblk}

\graphicspath{{media/}}     

\title{AROS: Affordance Recognition with One-Shot Human Stances
}

\author[1]{Abel Pacheco-Ortega}
\author[1,2]{Walterio Mayol-Cuevas}

\affil[1]{Visual Information Lab, Department of Computer Science, University of Bristol, UK 

            \texttt{\{abel.pachecoortega, walterio.mayol-cuevas\}@bristol.ac.uk} }
\affil[2]{Amazon, Seattle, Washington, USA}

\begin{document}
\maketitle

\begin{abstract}
We present AROS, a one-shot learning approach that uses an explicit representation of interactions between highly-articulated human poses and 3D scenes. The approach is one-shot as the method does not require re-training to add new affordance instances. Furthermore, only one or a small handful of examples of the target pose are needed to describe the interaction. Given a 3D mesh of a previously unseen scene, we can predict affordance locations that support the interactions and generate corresponding articulated 3D human bodies around them. We evaluate on three public datasets of scans of real environments with varied degrees of noise. Via rigorous statistical analysis of crowdsourced evaluations, results show that our one-shot approach outperforms data-intensive baselines by up to 80{\%}.
\end{abstract}

\keywords{affordances detection \and scene understanding \and human interactions \and visual perception \and affordances}

\section{Introduction}	\label{sec:intro}

Vision evolved to make inferences in a 3D world, and one of the most important assessments we can make is what can be done where.
Detecting such environmental affordances allows to identify locations that support actions such as stand-able, walk-able, place-able, sit-able, and so on. Human affordance detection is central in scene analysis and scene understanding, but also potentially in object detection and labeling (via their use) and eventually useful too for scene generation.

Recent approaches have worked towards providing such key competency to artificial systems via iterative methods such as deep learning \cite{Du_2020_CVPR,Zhang2020,9665916,Carion2020,Bochkovskiy2020}. The effectiveness of these data-driven efforts is highly dependent on the number of classes, the number of examples per class and their diversity. Usually, a dataset consists of thousands of examples, and the training process requires a significant amount of hand tuning and compute resources. When a new category needs to be incorporated,  further sufficient samples need to be provided, and training remade. The appeal for one-shot training methods is clear.

\begin{figure}[t]
    \centering
    \includegraphics[width=\linewidth]{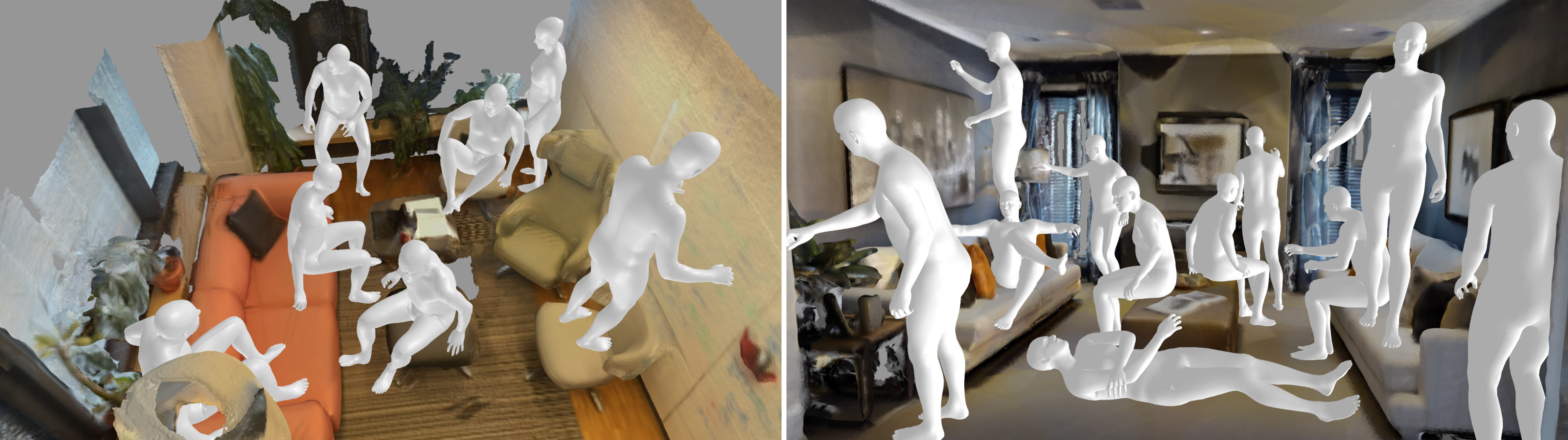}
    \caption{ 
        AROS is capable of detecting human-scene interactions with one-shot learning. Given a scene, our approach can detect locations that support interactions and hallucinate the interacting human body in a natural and plausible way. Images show examples of detected sit-able, reach-able, lie-able and stand-able locations.}
    \label{fig:human_scene_interactions}
\end{figure}

Often, human pose-in-scene detection is conflated with object detection or other semantic scene recognition. For example, training to detect sit-able locations through the recognition of chairs. This is a flawed approach for general action-scene understanding, first since people can find many non-chair locations where they can and do sit, e.g. on tables or cabinets (\Cref{fig:human_scene_interactions}). Second, an object-driven approach may fail to consider that affordance detection depends on the object pose and its surroundings ---it should not detect a chair as sit-able if it is upside-down or if an object is over it. Finally, object detectors alone may struggle to perceive a potentially sit-able place if not covered during training.

To address these limitations, AROS (Affordance Recognition with One-shot human Stances) uses a direct representation of human-scene affordances. It extracts an explainable geometrical description by analysing proximity zones and clearance space between the interacting entities. The approach allows training from one or very few data samples per affordance and is capable of handling noisy scene data as provided by real visual sensors such as RGBD and stereo cameras.

In summary, our contributions are: (1) a one-shot learning geometric-driven affordance descriptor that captures both proximity zones and the clearance space around the interaction. (2) We generate an improvement on 3D scene representation that permits reliable detections and generation of physically plausible human-scene interactions. (3) We validate results with a comprehensive statistical analysis and tests that show our approach generates natural and physically plausible human-scene interaction with better performance than intensively trained state of the art methods. (4) Our approach demonstrates control on the kind of human-scene interaction sought which permits exploring scenes with a concatenation of affordances.

\section{Related Work}

Following Gibson's suggestion that affordances are what we perceive when looking at scenes or objects \cite{Gibson1977}, the perception of human affordances with computational approaches has been extensively explored over the years. 
Before data-intensive approaches, Gupta et. al \cite{Gupta2011} used a environment geometric estimation and a voxelized discretization of 4 human poses to measure the environment affordances capabilities. This human pose method was employed by Fouhey et al. in \cite{Fouhey2015} to automatically generate thousands of labeled RGB frames from the NYUv2 dataset \cite{Silberman:ECCV12} for training a neural network and a set of local discriminative templates that permits the detection of 4 human affordances. A related approach was explored by Roy et al. \cite{Roy2016} where detection was performed for 5 different human affordances through a pipeline of CNNs that includes the extraction of mid-level cues trained on the NYUv2 dataset \cite{Silberman:ECCV12}. 
Luddecke et al. \cite{Luddecke2017} implemented a residual neural network for detecting 15 human affordances, trained using a look-up table that assigns affordances to object parts on the ADE20K dataset \cite{zhou2017scene}.


Another research line has been the creation of action maps. Savva et al. \cite{Savva2014} generate affordance maps by learning relations between human poses and geometries in recorded human actions. Piyathilaka et al. \cite{Piyathilaka2015},  use human model skeletons positioned on different locations in an environment to measure geometric features and determine support. In \cite{Rhinehart2016}, Rhinehart et al. use egocentric videos as well as scenes, objects and actions classifiers to build up the action maps.


There have been efforts to use functional reasoning for describing the purpose of elements in the environment that help to define them. 
Grabner et al. \cite{Grabner2011} designed  a geometric detector for sit-able objects like chairs, while further explorations performed in \cite{zhu2016inferring,wu2020chair} included physics engines to ponder also physics constrains like collision, inertia friction, and gravity.


Wang et al. \cite{wang2017binge} proposed an affordance predictor and a 2D human interaction generator trained on a more than 20K images extracted from sitcoms with and without humans interacting with the environment. Li et al. \cite{li2019putting} extend of this work by developing a 3D human pose synthesizer that learns on the same dataset of images but generates human interactions into input scenes represented as RGB, RGBD or depth images.

Jiang et al. \cite{Jiang2016} exploited the spatial correlation between elements and humans interactions on RGBD images to generate human interactions and improving object labeling. These methods use human skeletons for hallucinating body-environment configurations which reduce their representativeness as contacts, collisions and naturalness of the interactions can not be evaluated in a reliable manner. Ruiz and Mayol \cite{ruiz2020geometric} developed a geometric interaction descriptor for non-articulated, rigid object shapes. Given a 3D environment the method demonstrated good generalization on detecting physically feasible object-environment configurations. In the SMPLX human body representation \cite{pavlakos2019expressive},  Zhang et. al. \cite{Zhang_2020_CVPR} present a context-aware human body generator that learned on recordings from the PROX \cite{hassan2019resolving} dataset the distribution of 3D human poses conditioned to the scene depth and semantics. In an follow up effort,  Zhang et al. \cite{zhang2020place} developed a purely geometrical approach to model human-scene interactions by explicitly encoding the proximity between the body and the environment, thus only required a mesh as input.
Training CNNs and related data-driven methods require the use of most if not all of the labelled dataset, which e.g. in PROX \cite{hassan2019resolving}, there are 100K image frames.




\section{AROS}

Detecting human affordances in an environment is to find locations capable of supporting a given interaction between a human body and the environment. For example, the study of finding "suitable to sit" locations identifies all those places in which a human can sit, which can include a range of object "classes" (sofa, bed, chair, table, etc.). Our method is motivated to develop a descriptor that characterizes such general interaction through two key components and that is lightweight wrt data needs while outperforming alternative baselines.

These two components weight the extraction of characteristics from areas where there is high (contact) and low (clearance) physical proximity respectively between the entities in interaction.

Importantly, the representation allows one-shot training per affordance which is desirable to improve training scalability. Furthermore, our approach is capable to describe and detect interactions between noisy data representations obtained with spatial visual sensors and highly articulated human poses.

\begin{figure}[t]
    \centering
    \includegraphics[width=\linewidth]{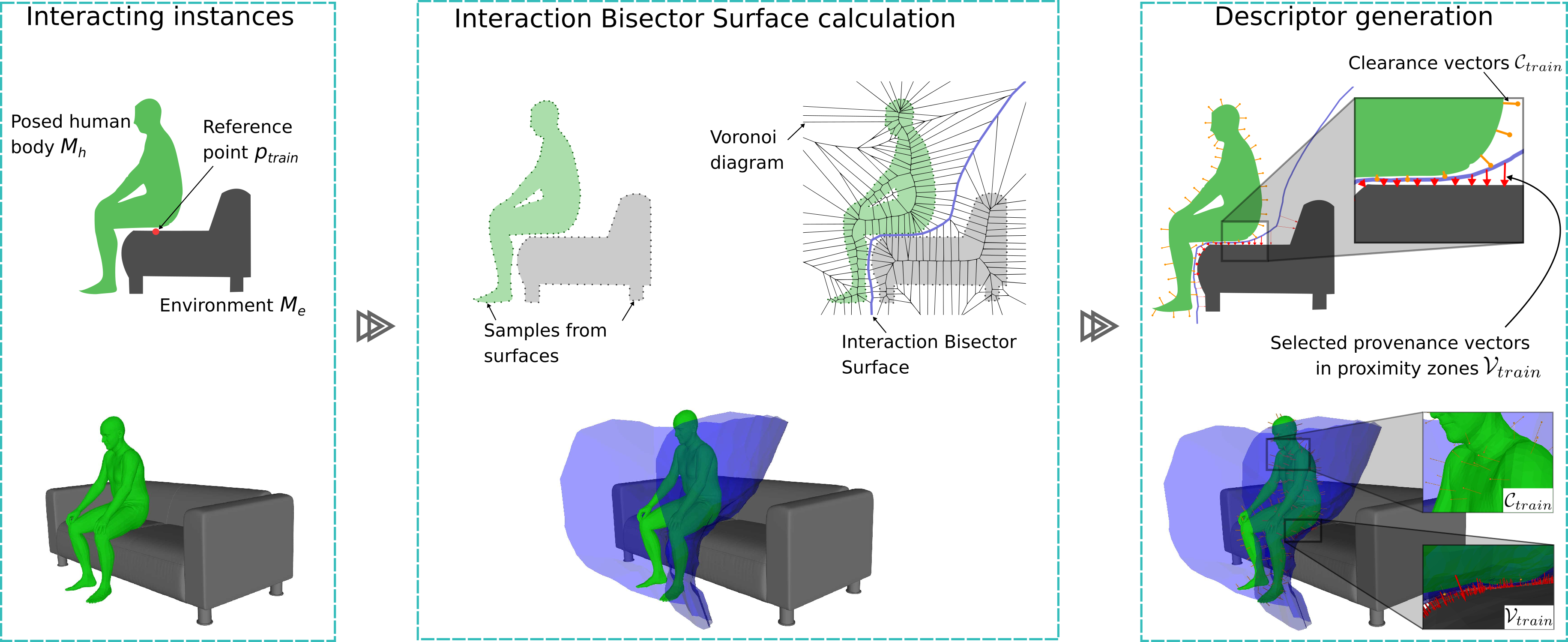}
    \caption{2D and 3D representations of our one-shot training pipeline. (Left) Posed human body $M_h$ interacting with an environment $M_e$ on a reference point $p_{train}$. (Center) Only during training, we calculate the Voronoi diagram with sample points from both the environment and the body surfaces, to generate an Interaction Bisector Surface (IBS). (Right) We use the IBS to characterise the proximity zones and the surrounding space with provenance and clearance vectors. A weighted sample of these provenance and clearance vectors, $\mathcal{V}_{train}$ and $\mathcal{C}_{train}$ respectively,  results in good generalisation of the interaction.}
    \label{fig:training}
\end{figure}

\subsection{A Spatial Descriptor for Spatial Interactions}\label{subsection:descriptor}
We are inspired by recent methods that have revisited geometric features such as the bisector surface for scene-object indexing \cite{Zhao2014} and affordance detection \cite{ruiz2020geometric}. Starting from a spatial representation makes sense if it helps to reduce data training needs and simplifies explainability ---as long as it can outperform data-intensive approaches.
Our affordance descriptor expands on the Interaction Bisector Surface (IBS) \cite{Zhao2014}, an approximation of the well-known Bisector Surface (BS) \cite{Peternell2000}. Given two surfaces $S_1, S_2 \in \mathbb{R}^3$, the BS is the set of sphere centres that touch both surfaces at one point each. Due to its stability and geometrical characteristics, the IBS has been used in context retrieval, interaction classification, and functionality analysis \cite{Zhao2014,Hu2015,Hu2016,Zhao2016,Zhao2017,ruiz2020geometric}. Our approach expands on these ideas and is geometrically intuitive and straightforward. It explicitly captures areas that are important to be in scene-contact and those that are not. Importantly, we show how this approach can generalize from just one or a small number of samples to a large unseen number of scenes. 

Our one-shot training process represents interactions by 3-tuples ($M_h$, $M_e$, $p_{train}$), where $M_h$ is a posed human body mesh, $M_e$ is an environment mesh and $p_{train}$ is the reference point on $M_e$ that supports the interaction.
Let $P_h$ and $P_e$ sets of samples on $M_h$ and $M_e$ respectively, their IBS $\mathcal{I}$ is defined as:

\begin{align}
    \mathcal{I} &= \big\{p \mid \min_{ p'_h \in P_h} \lVert p-p'\rVert=\min_{ p'_e \in P_e} \lVert p-p'\rVert \big\} \label{eq:ibs_definition} 
\end{align}

We use the Voronoi diagram $\mathcal{D}$ generated with $P_h$ and $P_e$ to produce $\mathcal{I}$. By construction, every ridge in $\mathcal{D}$ is equidistant to the couple of points that defined it. Then, $\mathcal{I}$ is composed of ridges in $\mathcal{D}$ generated because of points from both $P_h$ and $P_e$. An IBS can reach infinity, but we limit $\mathcal{I}$ by clipping it with the bounding sphere of $M_h$ with tolerance ${ibs}_{rf}$.

The number and distribution of samples in $P_h$ and $P_e$ are crucial for a well constructed discrete IBS. A low rate of sampled points degenerates on an IBS that pierces boundaries of $M_h$ or $M_e$. A higher density is critical in those zones where the proximity is small. To populate $P_h$ and $P_e$, we first use a Poisson disc sampling strategy \cite{Yuksel2015} to generate $ibs_{ini}$ evenly distributed samples on each mesh surface. Then, we perform a \emph{counter-part sampling} that increases the numbers of samples in $P_e$ by including the closest points on $M_e$ to elements in $P_h$, and similarly, we incorporate in $P_h$ the closest point on $M_h$ to samples in $P_e$. We perform the \emph{counter-part sampling} strategy $ibs_{cs}$ times to generate a new $\mathcal{I}$.
However, we have found that for intricate human-scene poses, convergence to an IBS without mesh piercing is challenging. If the IBS is penetrating the scene, we perform a \emph{collision point sampling} strategy. This adds as sampling points, a sub sample of points where collisions happen and their counterpart points (body or environment). We then simply re-compute the IBS and repeat the \textit{counter-part sampling} and \textit{collision point sampling} strategies until we find a candidate $\mathcal{I}$ that does not collide with $M_h$ nor $M_e$. This is a straightforward process that can be implemented efficiently.

To capture the regions of interaction proximity on our enhanced IBS as per above, we use the notion of provenance vectors \cite{ruiz2020geometric}. The \textit{provenance vectors} of an interaction start from any point on $\mathcal{I}$ and finish on $M_e$. Formally:

\begin{align}
V_p = \big\{(a, \vec{v}) \mid a \in \mathcal{I}  ,\ \vec{v} = \underset{e \in M_e}{\arg\min} \lVert e - a \rVert - a \big\}
\end{align}
where $a$ is the stating point of the delta vector $\vec{v}$ to the nearest point on $M_e$.

\emph{Provenance vectors} inform about the direction and distance of the interaction; the smaller $|\vec{v}|$ the more importance in the description. Let $V'_p \subset V_p$ the subset of \textit{provenance vectors} that finish on any point in $P_e$, we perform a weighted randomized selection sampling of elements from $V'_p$ with the allocation of weights as follows:

\begin{equation}
w_i = 1-\frac{\lvert\vec{v}_i\rvert - \lvert\vec{v}_{min}\rvert}{\lvert\vec{v}_{max}\rvert - \lvert\vec{v}_{min}\rvert}
\ \ \ ,\  i=1,\ 2,\  \dots,\ |P_e|
\end{equation}
where $\lvert\vec{v}_{max}\rvert$ and $\lvert\vec{v}_{min}\rvert$ are the norms of the biggest and smallest vectors in $V'_p$ respectively. The selected \textit{provenance vectors} $\mathcal{V}_{train}$ integrate to our affordance descriptor with an adjustment to normalize their positions with the defined reference point $p_{train}$:

\begin{align}
    \mathcal{V}_{train} = \big\{(a'_i, \vec{v}_i) \mid  a'_i = a_i - p_{train} \ \ , \ i=  1,\ 2,\ \dots,\ num_{pv}  \big\}
\end{align}
where $num_{pv}$ is the number of samples from $V'_p$ to integrate. The \textit{provenance vectors} however, are on their own, insufficient to work well on highly articulated objects like human poses. They are unable to capture the whole nature of the interaction. We expand this concept by taking a more comprehensive description that considers both areas of the IBS that are proximal to surfaces and those that are not.

We include into our descriptor a set of vectors to define the clearance space necessary for performing the given interaction. Given $S_h$ an evenly sampled set of $num_{cv}$ points on $M_h$, the \textit{clearance vectors} that integrates to our descriptor $\mathcal{C}_{train}$ on the interaction are defined as follows:

\begin{align}
    \mathcal{C}_{train} = \big\{(s'_j, \vec{c}_j) \mid   s'_j=s_j - p_{train}\ ,\ s_j \in S_h\ ,\ \vec{c}_j = \psi(s_j,\ \hat{n}_j,\ \mathcal{I}) \big\} \\
    \psi(s'_j, \hat{n}_j, \mathcal{I}) = \begin{cases}
        d_{max} \cdot \hat{n}_j                         \ \ \ & \text{if\ \ } \varphi(s_j,\ \hat{n}_j,\ \mathcal{I})>d_{max}\\
        \varphi(s_j, \hat{n}_j, \mathcal{I}) \cdot \hat{n}_j   \ \ \ & \text{otherwise}
    \end{cases}
\end{align}
where $p_{train}$ is the defined reference point,  $\hat{n}_i$ is the unit surface normal vector on sample $s_j$, $d_{max}$ is the maximum norm of any $\vec{c}_j$, and $\varphi(s_j,\ \hat{n}_j,\ \mathcal{I})$ is the distance traveled by a ray with origin $s_j$ and direction $\hat{n}_i$ until collision with $\mathcal{I}$.

Formally, our affordance descriptor, \textbf{AROS}, is defined as:

\begin{align}
    f:(M_h, M_e, p_{train}) \longrightarrow (\mathcal{V}_{train},\mathcal{C}_{train}, \hat{n}_{train} )
\end{align}
where $\hat{n}_{train}$ is the unit normal vector on $M_e$ at $p_{train}$. We calculate $\hat{n}_{train}$ for speeding up the detection process.

\subsection{Human Affordance Detection} 

\begin{figure}[t]
    \centering
    \includegraphics[width=\linewidth]{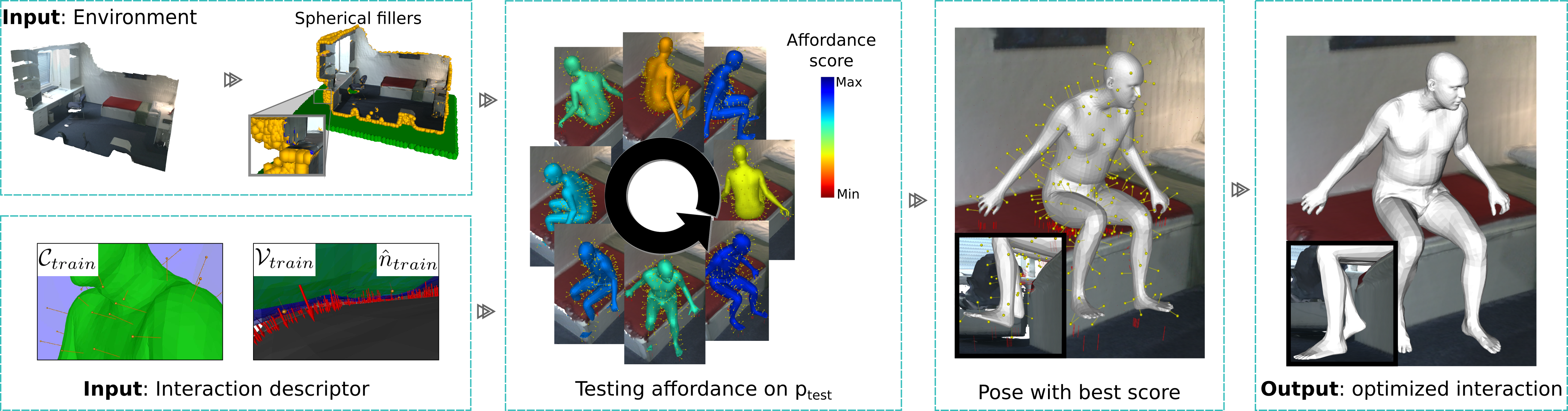}
    \caption{Approach for detecting human affordances. To mitigate 3D scan noise, the scene is augmented with spherical fillers for detecting collisions and SDF values. Our method detects if a test point in the environment can support an interaction by translating the descriptor to the test position over different orientations and measuring its alignment and collision rate. Then, the best-scored configuration is optimized to generate a more natural and physically plausible interaction with the environment. }
    \label{fig:testing}
\end{figure}

Let $\mathcal{A}=(\mathcal{V}_{train}, \mathcal{C}_{train}, \hat{n}_{train} )$ be an affordance descriptor, we define its rigid transformation with $\tau \in \mathbb{R}^3$ being a translation vector and $\phi$ being the rotation around $z$ defined by $R_\phi$.

Given a point $p_{test}$ on an environment mesh $M_{test}$ and its unit surface normal vector $\hat{n}_{test}$,  we determine that such location supports a trained interaction  $\mathcal{A}$ if we can find that (1) there is a small angle difference between $\hat{n}_{test}$ and $\hat{n}_{train}$, (2) once translated to $p_{test}$ and oriented with $\phi_{test}$, there is a correct alignment of $\mathcal{V}^{A}_{\phi\tau}$ and (3) a gated number of the $\mathcal{C}^{A}_{\phi\tau}$ are in collision with $M_{test}$.

A significant angle difference between $\hat{n}_{test}$ and $\hat{n}_{train}$ permits to short-cut the test and reject $p_{test}$ wrt $\mathcal{A}$. We establish  $\rho_{\vec{n}}$ as the decision threshold for the angle difference. $\rho_{\vec{n}}$ is adjustable based on the level of mesh noise.

If we find normals' match between $\hat{p}_{train}$ and $p_{test}$ vectors, we perform transformations over the interaction descriptor $\mathcal{A}$ with  $\tau=p_{test}$ and $n_{\phi}$ different $\phi=\phi_{test}$ values within $[0,2\pi]$. 
Per each 3-tuple $(\mathcal{V}^{A}_{\phi\tau},\ \mathcal{C}^{A}_{\phi\tau},\ \hat{n}_{train} )$ calculated we generated a set of rays $R_{pv}$ defined as follows:

\begin{align}
    R_{pv}=\Big\{(a''_i, \hat{\nu}_i)\ |\  \hat\nu_i = \frac{\vec{v}{i}}{ \|\vec{v}{i}\| } \ ,\  (a''_i, \vec{v}_i) \in \mathcal{V}^{A}_{\phi\tau}    \Big\}
\end{align}
where $a''_i$ is the starting point and $\hat{\nu}_i \in \mathbb{R}^3$ the direction of each ray. We extend each ray in $R_{pv}$ by $\epsilon^{pv}_i$ until collision with $M_{test}$ as

\begin{align}
    (a''+\epsilon^{pv}_i \cdot \hat{\nu}_i) \in M_{test}\ \ \ ,  \ \ i=1,2,\dots,num_{pv}
\end{align}
and comparing with the magnitude of each correspondent provenance vector in $\mathcal{V}^{A}_{\phi\tau}$. When any element in $R_{pv}$ extends further than a predetermined limit $max_{long}$, the collision with the environment is classified as non-colliding. We calculate the alignment score $\kappa$ as a sum differences between extended rays and \textit{provenance vectors} with

\begin{align}
    \kappa = \sum_{\forall i | \epsilon^{pv}_i \leq max_{long}} |\epsilon^{pv}_i- \vec{v}_i | 
\end{align}

The bigger $\kappa$ value, the less the support for the interaction on the $p_{test}$. We experimentally determine interaction-wise thresholds for the sum of differences $max_\kappa$ and the number of missing ray collision $max_{missings}$ that permit us to score the affordance capabilities on $p_{test}$.

\textit{Clearance vectors} are meant to fast-detect collision configurations by ray-mesh intersection calculation. Similar to \textit{provenance vectors}, we generate a set of rays $R_{cv}$ which origins and directions follows are determined by $\mathcal{C}^{A}_{\phi\tau}$. We extend rays in $R_{cv}$ until collision with the environment and calculate its extension $\epsilon^{cv}_j$. Extended rays with $\epsilon^{cv}_j \leq \|\vec{c}_j\|$ are considered as possible collisions. In practice, we also track an interaction-wise threshold to refuse affordance due to collisions $max_{collisions}$.

A sparse distribution of clearance vectors on bi-dimensional noisy meshes in a 3D space results in collisions that are not detected by \textit{clearance vectors}. To improve, we enhance scenes with a set of \textit{spherical fillers} that pad the scene (see \Cref{fig:testing}(a)). More details in the supplemental material.

\subsubsection{Pose Optimization} \label{subsubsection:optimization_stage}
After a positive detection, we generate on the testing location, the body mesh representation as used on training. This generally has low levels of contact with the unseen environment. These gaps are because our descriptor base its construction on the bisector surface between the interacting entities. We could eliminate the gap by translating the body until touching the environment. But this na\"{\i}ve method generates configurations that visually lack naturalness, \Cref{fig:testing}(c).

Every human-environment configuration trained has an associated 3D human SMPL-X characterization that we keep and use to optimize the human pose as in \cite{zhang2020place} with the \textit{AdvOptim} loss function, using the SDF values pre-calculated in each scene with a grid of 256x256x256 positions.

\begin{figure}[t]
    \centering
    \includegraphics[width=\linewidth]{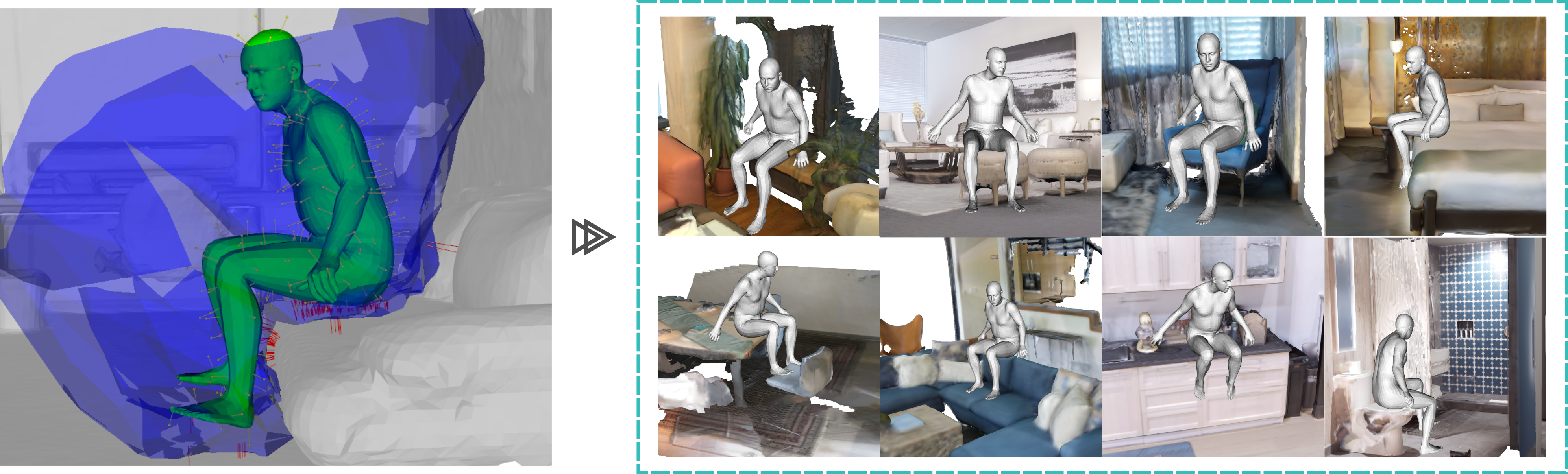}
    \caption{Our one-shot learning approach generalizes well on affordance detection. Only one example of an interaction to generate an AROS descriptor that generalizes well for detection of affordances over previously unseen environments.}
    \label{fig:same_iter_detection}
\end{figure}

Overall, we train a human interaction by generating its AROS descriptor from a single example, keeping the associated SMPL-X parameters of the body pose, and defining the contact regions that the body has with the environment. After a positive detection with AROS, we use the associated SMPL-X body parameters and its contact regions to close the environment-body gap and generate a more natural body pose as shown in \Cref{fig:testing}(Ouput). Our approach generalizes well on description of interaction and generates natural and physical plausible body-environment configurations over novel environments with just one example for training (see \Cref{fig:same_iter_detection}).

\section{Experiments}
We conduct experiments in various environment configurations to examine the effectiveness and usefulness of the affordance recognition performed by AROS. Our experiments include several perceptual studies, as well as a \textit{physical plausibility} evaluation of the body-environment configurations generated.

\textbf{Datasets.} The PROX dataset (\cite{hassan2019resolving}) includes data from 20 recordings of subjects interacting within 12 scanned indoor environments. An SMPL-X body model (\cite{pavlakos2019expressive}) is used to characterize the shape and pose of the human within each frame in recordings. Following the setup in \cite{zhang2020place}, we use the rooms MPH16, MPH1Library, N0SittingBooth and N3OpenArea for testing purposes and train on data from other PROX scenes. We also perform evaluations on 7 scanned scenes from the MP3D dataset (\cite{Matterport3D}) as well as on 5 scenes from the Replica dataset (\cite{replica19arxiv}). We calculate the \textit{spherical fillers} and SDF values of all 3D scanned environments.

\textbf{Training.} We manually select 23 frames in which subjects interact in one of the following ways: sitting, standing, lying down, walking, or reaching. From these selected human-scene interactions, we generate the AROS descriptors and retain the SMPLX parameters associated with the human poses. 

To generate the IBS surface associated with each trained interaction, we use an initial sampling set of $ibs_{ini}=400$ on each surface, execute the \textit{counter-part sampling} strategy $ibs_{cs}=4$ times, and crop the generated IBS surface $\mathcal{I}$ with $ibs_{rf}=1.2$. The AROS descriptors are a compound of $num_{pv}=512$ \textit{provenance vectors} and $num_{cv}=256$ \textit{clearance vectors} that extend up to $d_{max}=5[cm]$ each.

The interaction-wise thresholds $max_\kappa$, $max_{missings}$, and $max_{collisions}$ are established experimentally, and $max_{long}$ is 1.2 times the radius of the sphere used to crop $\mathcal{I}$. We use a moderate angle difference threshold of $\rho_{\vec{n}} = \pi/3$, in $n_{\phi}=8 $ different directions.

With 512 provenance vectors $\mathcal{V}_{train}$, and 256 clearance vectors $\mathcal{C}_{train}$, the AROS descriptor characterizes an interaction with less than 40KB.

\textbf{Baselines.} We compare our approach with the state-of-the-art PLACE (\cite{zhang2020place}) and POSA (contact only) (\cite{hassan2021populating}).  PLACE is a pure scene-centric method that only requires a reference point on an scanned environment to generate a body human performing around it. However, PLACE does not have control over the type of interaction detected/generated. We used naive (PLACE) and optimized versions of this approach in experiments (PLACE, PLACE SimOptim, and PLACE AdvOptim). POSA is a human-centric approach that, given a posed human body mesh, calculates the zones on the body where contact with the scene would occur and uses this feature map to place the body in the environment. We encourage a fair comparison by evaluating the naive and optimized POSA version that considers only contact information and excludes semantic information (POSA and POSA optimized). In our studies, POSA was executed with the same human shapes and poses used to train AROS.


\subsection{Physical plausibility}\label{subsection:physical_plausibility}

We evaluate the physical plausibility of the compared approaches mainly following \cite{zhang2020place,Zhang_2020_CVPR}.  Given the SDF values of a scene and a body mesh generated: a) the \textit{contact score} is assigned to 1 if any mesh vertex has a negative SDF value and is evaluated as 0 otherwise, b) the \textit{non-collision score} is the ratio of vertices with a positive SDF value, and c)in order to measure the severity of the body-environment collision on positive contact, we include the \textit{collision-depth score}, which averages the depth of the collisions between the scene and the generated body mesh. 


\subsubsection{Ablation Study}\label{subsection:ablation_study}
We evaluate the influence of \textit{clearance vectors}, spherical fillers, and different optimizers on the PROX dataset. Three different optimization procedures are evaluated. The \textit{downward} optimizer translates the generated body downward (-Z direction) until it contacts the environment. The ICP optimizer uses the well-known Interactive Closest Point algorithm to align the body vertices with the environment mesh. The AdvOptim optimizer is described in \Cref{subsubsection:optimization_stage}.

As shown in \Cref{tab:ablation_study}, models without \textit{clearance vectors} got the highest collision-depth scores on models with the same optimizer. AROS models present a reduction in contact and collision-depth scores in all cases that consider \textit{clearance vectors} in their descriptors to avoid collision with the environment. Spherical fillers have a great influence on avoiding collisions, producing the best scores in all metrics per optimizer. \textit{ICP} optimizer closes the body-environment gaps but drastically reduces performance on both collision scores while the \textit{AdvOptim} and \textit{downward} optimizers keep a trade-off between collision and contact. The best performance is achieved with affordance descriptors composed of \textit{provenance} and \textit{clearance vectors}, tested in scanned environments enhanced with \textit{spherical fillers}, and where interactions optimized with the \textit{AdvOpt} optimizer.

\begin{table}[t]
    \centering
    \caption{Ablation study evaluation scores ($^\uparrow$: benefit, $^\downarrow$: cost). Best trade-off of scores per optimizer in boldface.}
    \label{tab:ablation_study}
        \begin{tabular}{@{}>{\raggedright\arraybackslash}m{0.14\linewidth}
                >{\centering\arraybackslash}m{0.10\linewidth}
                >{\centering\arraybackslash}m{0.11\linewidth}
                *{5}{>{\centering\arraybackslash}m{0.09\linewidth}} 
                @{}}
            \toprule
            Descriptor integrated by  & Spherical fillers & Optimizer               & non collision$^\uparrow$ & contact$^\uparrow$ & collision depth$^\downarrow$     \\ \midrule
        	$\mathcal{V}_{train}$      & no                & \multirow{3}*{w/o}      & 0.9348                   & 0.7998             & 1.4132                           \\
        	$\mathcal{V}_{train},\mathcal{C}_{train}$               & no                &                         & 0.9504                   & 0.6901             & 0.6757                           \\
        	$\mathcal{V}_{train},\mathcal{C}_{train}$               & yes               &                         & \textbf{0.9623}          & \textbf{0.5448}    & \textbf{0.1573}                  \\ \midrule
        	$\mathcal{V}_{train}$      & no                & \multirow{3}*{ICP}      & 0.5820                   & 1.0000             & 7.3770                  \\
        	$\mathcal{V}_{train},\mathcal{C}_{train}$               & no                &                         & 0.5775                   & 1.0000             & 7.2180                           \\
        	$\mathcal{V}_{train},\mathcal{C}_{train}$               & yes               &                         & \textbf{0.6299}          & 1.0000             & \textbf{6.2665}                  \\ \midrule
        	$\mathcal{V}_{train}$      & no                & \multirow{3}*{downward} & 0.9271                   & 0.9377             & 1.4380                           \\
        	$\mathcal{V}_{train},\mathcal{C}_{train}$                & no                &                         & 0.9496                   & 0.9036             & 0.7089                           \\
        	$\mathcal{V}_{train},\mathcal{C}_{train}$               & yes               &                         & \textbf{0.9641}          & \textbf{0.8603}    & \textbf{0.1807}                  \\ \midrule
            $\mathcal{V}_{train}$      & no                & \multirow{3}*{AdvOptim} & 0.9552                   & 0.9638             & 2.0249                           \\
        	$\mathcal{V}_{train},\mathcal{C}_{train}$               & no                &                         & 0.9717                   & 0.9508             & 1.2325                           \\
        	$\mathcal{V}_{train},\mathcal{C}_{train}$               & yes               &                         & \textbf{0.9818}          & \textbf{0.9403}    & \textbf{0.6341}                  \\ \bottomrule
        	\multicolumn{6}{r}{\scriptsize{Note: ICP stands for Interative Closest Point }} \\ 
        \end{tabular}
\end{table}




\subsubsection{Comparison with the state of the art}
We generate 1300 interacting bodies per model in each scene and report the averages of calculated non-collision, contact and collision-depth scores. The results are presented in \Cref{tab:physical_plausibility}.  In all datasets, interacting bodies generated with our approach present a good trade-off with high non-collision but low contact and collision-depth scores. 

\begin{table}[t]
    \centering
    \caption{Physical plausibility. Non-collision, contact and collision-depth scores ($^\uparrow$: benefit, $^\downarrow$: cost) before and after optimization. Best results boldface} 
    \label{tab:physical_plausibility}
    \resizebox{\textwidth}{!}{%
        \scriptsize{
            \begin{tabular}{@{}
                    *{2}{>{\raggedright\arraybackslash}m{0.10\linewidth}}
                    *{9}{>{\centering\arraybackslash}m{0.07\linewidth}}@{}
                }
                \toprule
                && \multicolumn{3}{c}{non collision$^\uparrow$} & \multicolumn{3}{c}{contact$^\uparrow$} & \multicolumn{3}{c}{collision depth$^\downarrow$} \\ \cmidrule(lr){3-5} \cmidrule(lr){6-8}\cmidrule(lr){9-11}
                Model      &Optimizer& PROX   & MP3D   & Replica         & PROX   & MP3D   & Replica   & PROX   & MP3D   & Replica           \\ \midrule
                PLACE               &w/o& 0.9207 & 0.9625 & 0.9554          & 0.9125 & 0.5116 & 0.8115    & 1.6285          & 0.8958 & 1.2031            \\
                PLACE       &SimOptim& 0.9253 & 0.9628 & 0.9562          & 0.9263 & 0.5910 & 0.8571    & 1.8169          & 1.0960 & 1.5485            \\
                PLACE       &AdvOptim& 0.9665 & 0.9798 & 0.9659          & 0.9725 & 0.5810 & 0.9931    & 1.6327          & 1.1346 & 1.6145            \\
                POSA        & w/o  & \textbf{0.9820} & 0.9792 & 0.9814& 0.9396 & 0.9526 & 0.9888    & 1.1252 & 1.5416 & 2.0620            \\
                POSA        & optimized  &0.9753 & 0.9725 & 0.9765    & \textbf{0.9927} & \textbf{0.9988}& \textbf{0.9963}  & 1.5343  & 2.0063 &  2.4518           \\
                AROS         &w/o& 0.9615 & \textbf{0.9853} & \textbf{0.9931} & 0.5654 & 0.3287 & 0.4860    & \textbf{0.1648} & \textbf{0.1326} & \textbf{0.2096} \\
                AROS    &AdvOptim& 0.9816 & \textbf{0.9853} & 0.9883 & 0.9363 & 0.6213 & 0.8682 & 0.6330 & 0.8716 & 0.8615\\
                \bottomrule
            \end{tabular}
         }
    }
\end{table}

\subsection{Perception of Naturalness}
We use Amazon Mechanical Turk to compare and evaluate the naturalness of body-environment configurations generated by our approach and baselines. We use only the best version of the compared methods (with optimizer). Each scene in our test set was used equally to select 162 locations around which the compared approaches generate human interactions. MTurk judges observed all human-environment pairs generated through dynamic views, allowing us to showcase them from different perspectives. Each judge performed 11 random-selected assessments, without repetition,  that included two control questions to detect and exclude untrustworthy evaluators. Three different judges accomplished each of the evaluations. Our perceptual experiments include individual and comparison studies for each comparative carried out.

In our \textbf{side by side comparison studies}, interactions detected/generated from two approaches are exposed simultaneously. Then, MTurkers were asked to respond to the question "Which example is more natural?" by direct selection. 

We used the same set of interactions for the \textbf{individual evaluation studies}, where judges rated every individual human-scene interaction by responding to "The human is interacting very naturally with the scene. What is your opinion?" with a 5-point Likert scale according to its agreement level : 1. strongly disagree, 2. disagree, 3. neither disagree nor agree, 4. agree, and 5. strongly agree.

\subsubsection{Randomly selected test locations} \label{subsection:random_locations}

The first group of studies compares human-scene configurations generated at randomly selected locations. 
On the \textbf{side by side comparison study} that contrast AROS with PLACE, our approach was selected as more natural in \% 60.7 of all assessments. Compared to POSA, ours is selected in \%72.6 of all tests performed. The results per dataset are presented in \Cref{tab:comparison_study}/left.

\textbf{Individual evaluation studies} also suggest that AROS produce more natural interactions (see \cref{tab:contingency_table_challenging_points}).  The mean and standard deviations of these scores given by the judges to PLACE are 3.23$\pm$1.35 in comparison with AROS 3.39$\pm$1.25  while in the second study these statistics obtained by  POSA were 2.79$\pm$1.18 in contrast with  AROS 3.20$\pm$1.18. Evaluation scores of AROS have a larger mean and a narrower standard deviation compared to baselines. However, these descriptive statistics should be cautiously used as evidence to determine a performance difference because it assumes that the distribution of scores approximately resemble a normal distribution and that the ordinal variable was perceived as numerically equidistant by judges. Regrettably, Shapiro-Wilk tests (\cite{shapiro1965analysis}) performed on data show that the score distributions depart from normality in both evaluation studies, PLACE/AROS and POSA/AROS with $p<0.01$. 

Based on this, we performed a chi-square test of homogeneity (\cite{franke2012chi}) with significance level $\alpha=0.05$, to determine if the distributions of evaluation scores are statistically similar. If we found significance, the level of association between the approach and the distribution of the scores was determined by calculating the Cramer's V value ($V$) (\cite{cramer1946mathematical}). 

Data from the PLACE/AROS evaluation suggest that there is no statistically significant difference between score distributions ($\chi_{(4)}^2=9.34$, $p=0.053$). A larger sample size is necessary to observe statistical significance with a minor or negligible size effect, nonetheless, findings could not be substantially interesting due to even minute differences will turn on. However, data from the POSA/AROS evaluation study showed that our approach performs better than POSA ($\chi_{(4)}^2=32.33$, $p<0.001$) with a medium level of association ($V=0.1823$).


\begin{table}[t]
    \centering
    \caption{Cross-tabulation data of individual evaluation studies on randomly selected locations. Best in boldface.}	
    \label{tab:contingency_table_random_points_all_datasets}
    \resizebox{\textwidth}{!}{%
    \begin{tabular}{l:llccccc} 
    \toprule
    \multicolumn{1}{l}{\begin{tabular}[c]{@{}l@{}}Individual \\evaluation study\end{tabular}} & Model &                 & \begin{tabular}[c]{@{}c@{}}1. strongly\\ disagree\end{tabular} & 2.disagree    & 3. neither & 4. agree      & \begin{tabular}[c]{@{}c@{}}5. strongly\\ agree\end{tabular}  \\ 
    \midrule
    \multirow{4}{*}{PLACE v AROS}                                                             & PLACE & Observed frequency         & 68                                                             & 98            & 70         & 153           & 97                                                           \\
                                                                                              &       & \% within model & 14.0                                                           & 20.2          & 14.4       & 31.5          & 19.9                                                         \\
                                                                                              & AROS  & Observed frequency         & \textbf{43}                                                    & \textbf{98}   & 64         & \textbf{187}  & \textbf{94}                                                  \\
                                                                                              &       & \% within model & \textbf{8.8}                                                   & \textbf{20.2} & 13.2       & \textbf{38.5} & \textbf{19.3}                                                \\ 
    \midrule
    \multirow{4}{*}{POSA v AROS}                                                              & POSA  & Observed frequency         & 64                                                             & 173           & 89         & 123           & 37                                                           \\
                                                                                              &       & \% within model & 13.2                                                           & 35.6          & 18.3       & 25.3          & 7.6                                                          \\
                                                                                              & AROS  & Observed frequency         & \textbf{29}                                                    & \textbf{136}  & 85         & \textbf{179}  & \textbf{57}                                                  \\
                                                                                              &       & \% within model & \textbf{6.0}                                                   & \textbf{28.0} & 17.5       & \textbf{36.8} & \textbf{11.7}                                                \\
    \bottomrule
    \end{tabular}
    }
\end{table}

\subsubsection{Challenging test locations} \label{subsection:challenging_locations}

\begin{figure}[t]
    \centering
    \includegraphics[width=\linewidth]{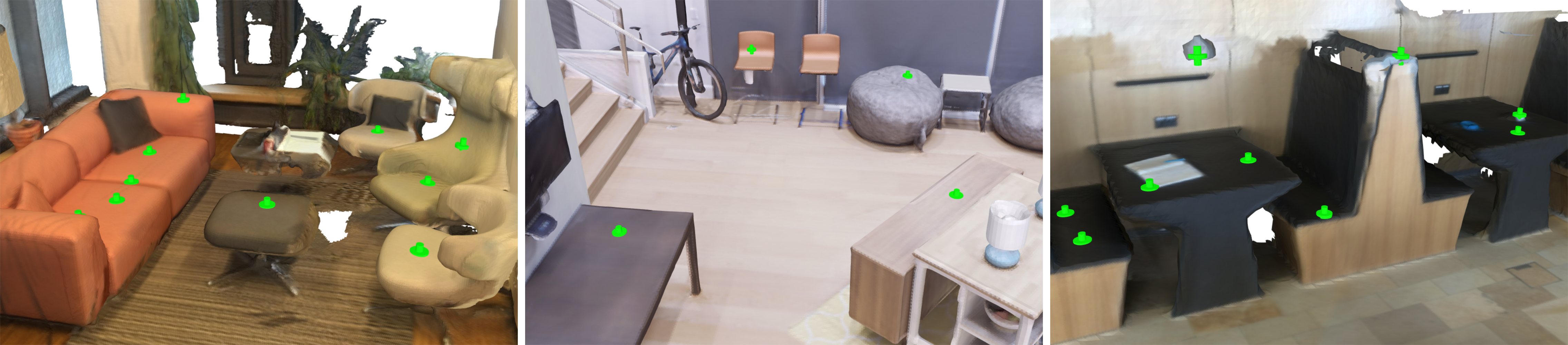}
    
    \caption{Selected by a golden annotator, green spots correspond to examples of meaningful, challenging locations for affordance detection.
    }
    \label{fig:challenging_positons}
\end{figure}

Our statistical analysis did not show any difference in the perception of naturalness of bodies generated by PLACE and AROS on randomly selected locations, but this sampling strategy frequently does not challenge the approaches. For example, a test could be oversimplified and inadequate for evaluations if the sampled scene has relatively large empty spaces where only the floor or a big plane surface surrounds the test locations. Therefore, we crowd-source the evaluations in a new set of more realistic locations provided by a golden-annotator (none of the authors) tasked with identifying areas of interest for human interactions \Cref{fig:challenging_positons}. These locations will be available for comparison.

The results of \textbf{side by side comparison studies} confirm that in \%60.6 of the comparisons with PLACE, AROS was considered more natural overall. Compared to POSA, AROS was marked with better performance in \%76.1 of all evaluations with a notorious difference in MP3D locations where AROS was evaluated as more natural in \%80.2 of the assesments (see \Cref{tab:comparison_study}/right for results per dataset).

\begin{table}[t]
    \centering
    \caption{MTurk side by side studies results on random and challenging locations. Best in boldface.}
    \label{tab:comparison_study}
        \begin{tabular}{llccclccc} 
        \toprule
        \multicolumn{1}{c}{} &  & \multicolumn{3}{c}{\begin{tabular}[c]{@{}c@{}}\% preferences in \\random locations\end{tabular}} &  & \multicolumn{3}{c}{\begin{tabular}[c]{@{}c@{}}\% preferences in\\challenging locations\end{tabular}}  \\ 
        \cmidrule{3-5}\cmidrule{7-9}
        Model  &  & MP3D          & PROX          & Replica     &  & MP3D          & PROX          & Replica\\ 
        \cmidrule{1-5}\cmidrule{7-9}
        PLACE  &  & 39.5          & 32.7          & 45.7        &  & 38.9          & 30.9          & 36.4 \\
        AROS   &  & \textbf{60.5} & \textbf{67.3} & \textbf{54.3}&  & \textbf{61.1} & \textbf{69.1} & \textbf{63.6}\\ 
        \midrule
        POSA   &  & 24.7          & 29.6          & 27.8         &  & 19.8          & 21.6          & 30.2\\
        AROS   &  & \textbf{75.3} & \textbf{70.4} & \textbf{72.2}&  & \textbf{80.2} & \textbf{78.4} & \textbf{69.8} \\
        \bottomrule
        \end{tabular}
\end{table}

As in the randomly selected test locations, a descriptive analysis of data from the \textbf{individual evaluation studies} on these new locations suggests that AROS performs better than other approaches with larger mean values and narrower standard deviations. The mean and standard deviation of the scores given by the judges to PLACE are 2.97$\pm$1.33 in comparison with AROS 3.44$\pm$1.19 while in the second study, these statistics obtained by POSA were 2.79$\pm$1.25 in contrast with AROS 3.5$\pm$1.25. However, a Shapiro-Wilk test performed on data shows that the score distributions also depart from normality with $p<0.01$ in both studies, PLACE/AROS and POSA/AROS,


A chi-square test of homogeneity, with $\alpha=0.05$ to determine whether both scores distributions are statistically similar on data from the PLACE/AROS evaluation study,  provides evidence that there is a difference in score distributions ($\chi_{(4)}^2=35.92$, $p<0.001$) with a medium level of association ($V=0.192$). 

However, an omnibus $\chi^2$ statistic does not provide information about the source of the difference between the score distributions. To this end, we performed a post-hoc analysis following the standardized residuals method described in \cite{agresti2018introduction}. To this end, as suggested in \cite{beasley1995multiple}, we corrected our significance level ($\alpha=0.05$) with the Sidak method (\cite{vsidak1967rectangular}) to its adjusted version $\alpha_{adj}=0.005$ with critical value $z=2.81$. The study revealed a significant difference in the qualification of the interactions generated by PLACE and AROS, with ours being qualified as natural more frequently. 

The residuals associated with AROS indicate with significant difference that the interactions generated by our approach were marked as "not natural" less frequently than expected: \emph{strongly disagree} ($z=-4.4, p <0.001$) and \emph{disagree} ($z=-2.98, p=0.002$).
Data also show a significant difference in favorable evaluations, where PLACE have less frequently positive evaluations than predicted by the hypothesis of independence in \emph{agree} ($z=-3.04, p<0.001$).  We also found a marginal significance, still in favor of AROS, in the frequency of \emph{strongly agree} evaluations ($z=-2.3, p=0.015$). 

No surprisingly, the chi-square test of homogeneity ($\alpha=0.05$ ) on data from the POSA/AROS evaluation study revealed that there is strong evidence of a difference in scores distributions ($\chi_{(4)}^2=75.13$, $p<0.001$) with a large level of association ($V=0.278$). The post-hoc analysis with standardized residuals concludes that the naturalness of human-scene interactions generated by AROS is, in the long term, better than those from POSA as well. \Cref{tab:contingency_table_challenging_points} shows the cross-tabulated data of the scores given by MTurkers and their standardized residual (critical value $z=2.81$ for $\alpha_{adj}=0.005$).

\begin{table}[h]
    \centering
    \caption{Cross-tabulation data of individual evaluation studies on challenging locations. A chi-square test of homogeneity on data provides evidence of difference in distribution of scores with $\alpha=0.05$. An analysis of residual indicates the source of such differences, an asterisk (*) indicates conservative statistical significance at $\alpha=0.05$, a double asterisk (**) denotes  statistical significance with $\alpha_{adj}=0.005$. Best in boldface.}
    \label{tab:contingency_table_challenging_points}
    \resizebox{\textwidth}{!}{%
       \begin{tabular}{l:llccccc} 
        \toprule
        \multicolumn{1}{l}{\begin{tabular}[c]{@{}l@{}}Individual \\evaluation study\end{tabular}} & Model &                       & \begin{tabular}[c]{@{}c@{}}1. strongly\\ disagree\end{tabular} & 2.disagree       & 3. neither & 4. agree        & \begin{tabular}[c]{@{}c@{}}5. strongly\\ agree\end{tabular}  \\ 
        \midrule
        \multirow{6}{*}{PLACE v AROS}                                                             & PLACE & Observed frequency     & 81                                                             & 131              & 54         & 161             & 59                                                           \\
                                                                                                  &       & \% within model       & 16.7\%                                                         & 27.0\%           & 11.1\%     & 33.1\%          & 12.1\%                                                       \\
                                                                                                  &       & Standardized residual & 4.44**                                                         & 2.98**           & -1.08      & -3.04**         & -2.43*                                                       \\
                                                                                                  & AROS  & Observed frequency     & \textbf{36}                                                    & \textbf{92}      & 65         & 207             & 86                                                           \\
                                                                                                  &       & \% within model       & \textbf{7.4\%}                                                 & \textbf{18.9\%}  & 13.4\%     & \textbf{42.6\%} & \textbf{17.7\%}                                              \\
                                                                                                  &       & Standardized residual & \textbf{-4.44**}                                               & \textbf{-2.98**} & 1.08       & \textbf{3.04**} & \textbf{2.43*}                                               \\ 
        \midrule
        \multirow{6}{*}{POSA v AROS}                                                              & POSA  & Observed frequency     & 86                                                             & 141              & 93         & 122             & 44                                                           \\
                                                                                                  &       & \% within model       & 17.7\%                                                         & 29.0\%           & 19.1\%     & 25.1\%          & 9.1\%                                                        \\
                                                                                                  &       & Standardized residual & 4.95**                                                         & 3.52**           & 1.70       & -2.88           & -6.57                                                        \\
                                                                                                  & AROS  & Observed frequency     & \textbf{35}                                                    & \textbf{94}      & 73         & \textbf{163}    & \textbf{121}                                                 \\
                                                                                                  &       & \% within model       & \textbf{7.2\%}                                                 & \textbf{19.3\%}  & 15.0\%     & \textbf{33.5\%} & \textbf{24.9\%}                                              \\
                                                                                                  &       & Standardized residual & \textbf{-4.95**}                                               & \textbf{-3.52**} & -1.70      & \textbf{2.88**} & \textbf{6.57**}                                              \\
        \bottomrule
        \end{tabular}
    }
\end{table}



\subsection{Qualitative results}
\begin{figure}[h]
    \centering
    \includegraphics[width=\linewidth]{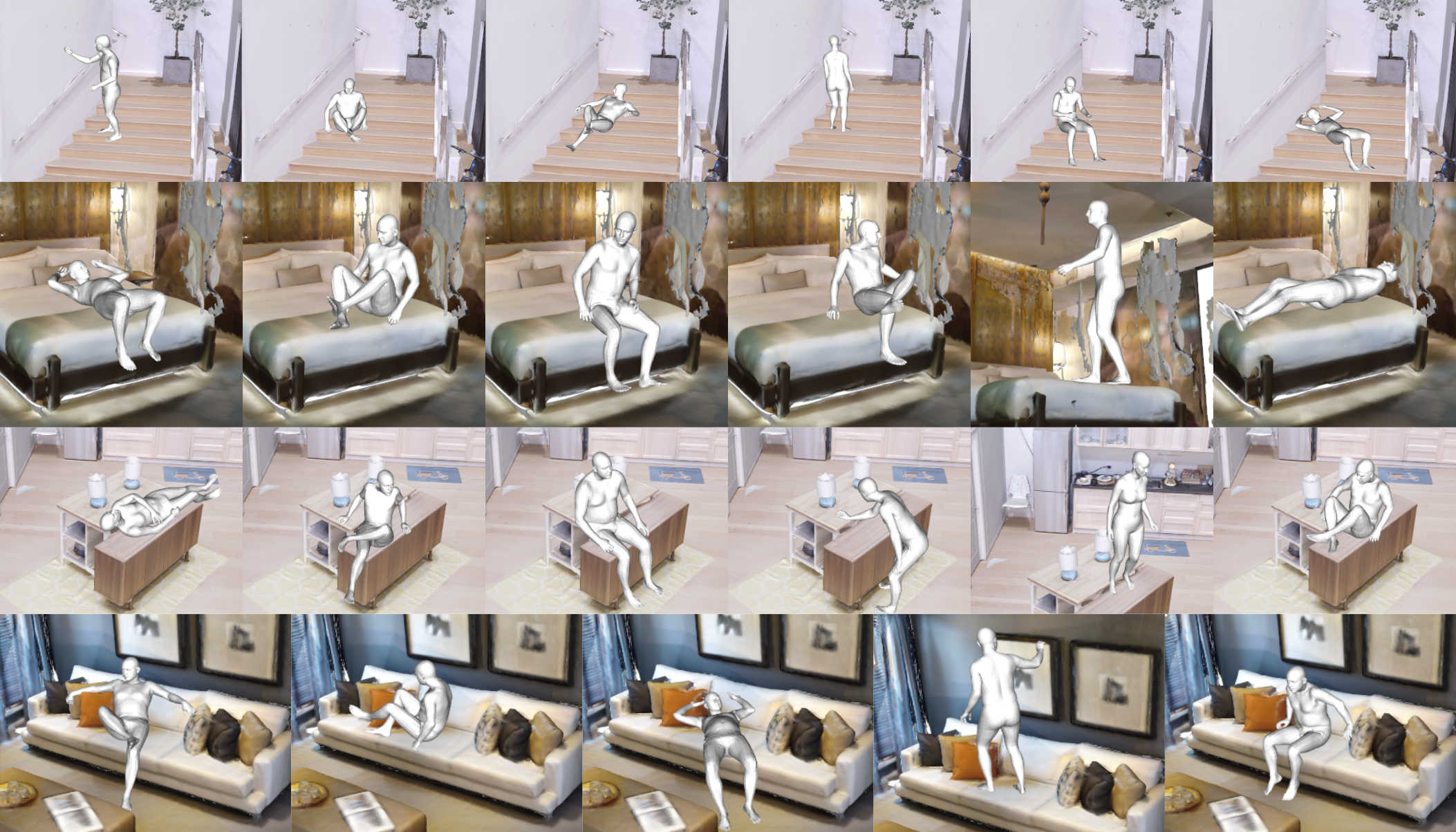}
    \caption{AROS shows good performance on a variety of novel scenes.}
    \label{fig:qualitative_results}
\end{figure}

\begin{figure}[h]
    \centering
    \includegraphics[width=\linewidth]{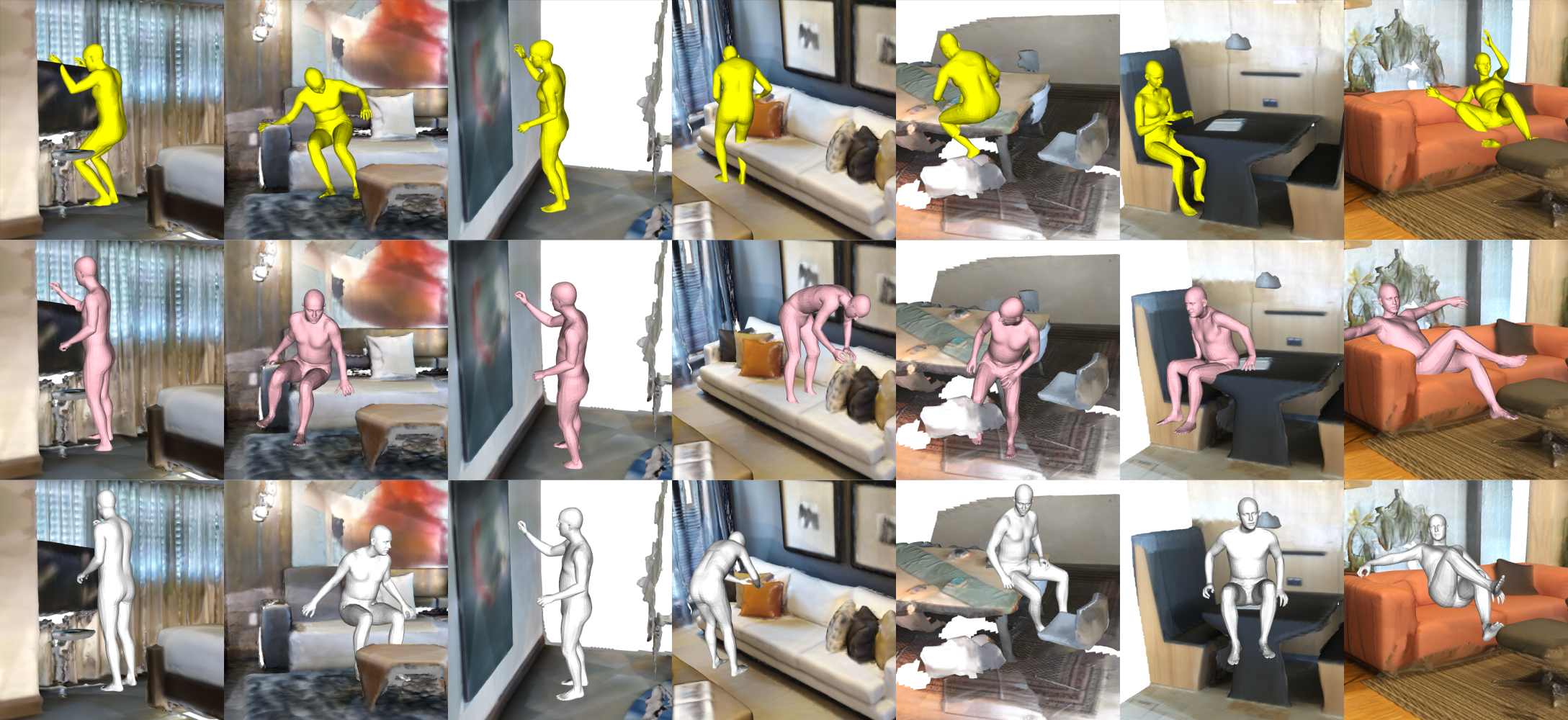}
    \caption{Qualitative challenging locations. PLACE (yellow), POSA (pink), AROS (silver).}
    \label{fig:place_POSAR_challenging_posiions}
\end{figure}

\begin{figure}[h]
    \centering
    \includegraphics[width=.8\linewidth]{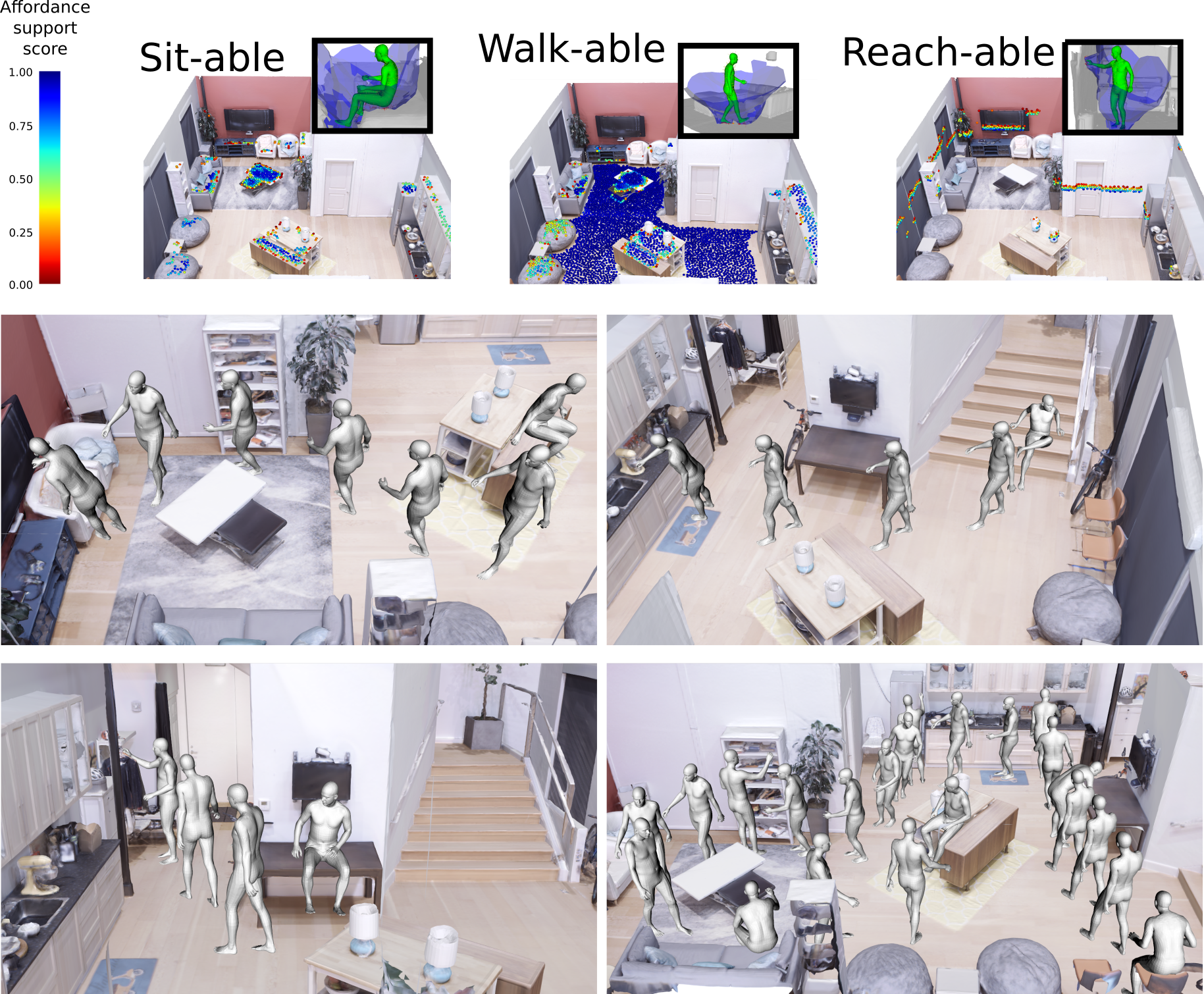}
    \caption{AROS can be used to create maps for action planning. Top: many locations in an environment are evaluated for three different affordances (sit-able, walk-able. reach-able). Bottom: AROS scores used to plan concatenated action milestones.}
    \label{fig:planning_pose_evolution}
\end{figure}

Experiments verify that our approach can realistically hallucinate human bodies that interact within a given environment in a natural and physically plausible manner. AROS allows us not only to determine the location on the environment in which we want the interaction to happen (the where), but also the specific type of interaction to be performed (the what).

The number and variety of interactions detected by AROS can easily be increased as a result of its one-shot training capacity. The more trained interaction, the more varied the human-scene configuration would be able to detect/hallucinate. \Cref{fig:qualitative_results} shows examples of different affordance detections around single locations.

AROS showed better performance in more realistic environment configurations where elements such as chairs, sofas, tables, and walls are presented and should be considered during the generation of body interactions. \Cref{fig:planning_pose_evolution} shows some examples of interaction generated by AROS and baselines over challenging locations.

Alternatively, affordances detections with AROS over several positions generate useful affordance maps for action planners (see \Cref{fig:planning_pose_evolution}).


\section{Conclusions}
In this work, we present AROS, a one-shot geometric driven affordances descriptor that builds on the bisector surface and combines proximity zones and clearance space to improve on affordance characterization. We introduce a generative framework that poses 3D human bodies interacting within a 3D environment in a natural and physically plausible manner. AROS shows a good generalization in unseen novel scenes. Furthermore, adding a new interaction to AROS is straightforward, since it needs as little as one example. Via rigorous statistical analysis, results show that our one-shot approach outperforms data-intensive baselines. Via rigorous statistical analysis, results show that our one-shot approach outperforms data-intensive baselines. We believe that explicit and interpretable description is valuable for complementing data-driven methods and open avenues for further work, including combining the strengths of both approaches.


\section*{Acknowledgments}
Abel Pacheco-Ortega thanks the Mexican Council for Science and Technology (CONACYT) for the scholarship provided for his postgraduate studies with the scholarship number 709908. Walterio Mayol-Cuevas thanks the visual egocentric research activity partially funded by UK EPSRC EP/N013964/1
\newpage
\clearpage
\bibliographystyle{unsrt}  
\bibliography{AROS}

\begin{thebibliography}{10}

\bibitem{Du_2020_CVPR}
Xianzhi Du, Tsung-Yi Lin, Pengchong Jin, Golnaz Ghiasi, Mingxing Tan, Yin Cui,
  Quoc~V Le, and Xiaodan Song.
\newblock {SpineNet: Learning Scale-Permuted Backbone for Recognition and
  Localization}.
\newblock In {\em Proceedings of the IEEE/CVF Conference on Computer Vision and
  Pattern Recognition (CVPR)}, jun 2020.

\bibitem{Zhang2020}
Hang Zhang, Chongruo Wu, Zhongyue Zhang, Yi~Zhu, Zhi Zhang, Haibin Lin, Yue
  Sun, Tong He, Jonas Mueller, R.~Manmatha, Mu~Li, and Alexander Smola.
\newblock {ResNeSt: Split-Attention Networks}.
\newblock apr 2020.

\bibitem{9665916}
Alexey Nekrasov, Jonas Schult, Or~Litany, Bastian Leibe, and Francis Engelmann.
\newblock {Mix3D: Out-of-Context Data Augmentation for 3D Scenes}.
\newblock In {\em 2021 International Conference on 3D Vision (3DV)}, pages
  116--125, 2021.

\bibitem{Carion2020}
Nicolas Carion, Francisco Massa, Gabriel Synnaeve, Nicolas Usunier, Alexander
  Kirillov, and Sergey Zagoruyko.
\newblock {End-to-End Object Detection with Transformers}.
\newblock In {\em European conference on computer vision}, pages 213--229.
  Springer, 2020.

\bibitem{Bochkovskiy2020}
Alexey Bochkovskiy, Chien-Yao Wang, and Hong-Yuan~Mark Liao.
\newblock {YOLOv4: Optimal Speed and Accuracy of Object Detection}.
\newblock apr 2020.

\bibitem{Gibson1977}
James~J. Gibson.
\newblock {The theory of affordances}.
\newblock In {\em Perceiving, acting and knowing. Toward and ecological
  psychology}. Lawrence Eribaum Associates, 1977.

\bibitem{Gupta2011}
Abhinav Gupta, Scott Satkin, Alexei~A. Efros, and Martial Hebert.
\newblock {From 3D scene geometry to human workspace}.
\newblock In {\em 2011 IEEE Conference on Computer Vision and Pattern
  Recognition (CVPR)}, pages 1961--1968. IEEE, jun 2011.

\bibitem{Fouhey2015}
David~F. Fouhey, Xiaolong Wang, and Abhinav Gupta.
\newblock {In Defense of the Direct Perception of Affordances}.
\newblock {\em arXiv preprint arXiv:1505.01085}, 2015.

\bibitem{Silberman:ECCV12}
Nathan Silberman, Derek Hoiem, Pushmeet Kohli, and Rob Fergus.
\newblock {Indoor segmentation and support inference from RGBD images}.
\newblock In {\em European conference on computer vision}, pages 746--760,
  2012.

\bibitem{Roy2016}
Anirban Roy and Sinisa Todorovic.
\newblock {A multi-scale CNN for affordance segmentation in RGB images}.
\newblock In {\em European conference on computer vision}, volume 9908 LNCS,
  pages 186--201. Springer, Cham, 2016.

\bibitem{Luddecke2017}
Timo Luddecke and Florentin Worgotter.
\newblock {Learning to Segment Affordances}.
\newblock In {\em The IEEE International Conference on Computer Vision (ICCV)
  Workshops}, pages 769--776. IEEE, oct 2017.

\bibitem{zhou2017scene}
Bolei Zhou, Hang Zhao, Xavier Puig, Sanja Fidler, Adela Barriuso, and Antonio
  Torralba.
\newblock {Scene Parsing through ADE20K Dataset}.
\newblock In {\em Proceedings of the IEEE Conference on Computer Vision and
  Pattern Recognition}, 2017.

\bibitem{Savva2014}
Manolis Savva, Angel~X. Chang, Pat Hanrahan, Matthew Fisher, and Matthias
  Nie{\ss}ner.
\newblock {SceneGrok: Inferring Action Maps in 3D Environments}.
\newblock {\em ACM transactions on graphics (TOG)}, 33(6):1--10, nov 2014.

\bibitem{Piyathilaka2015}
Lasitha Piyathilaka and Sarath Kodagoda.
\newblock {Affordance-map: Mapping human context in 3D scenes using
  cost-sensitive SVM and virtual human models}.
\newblock In {\em 2015 IEEE International Conference on Robotics and
  Biomimetics (ROBIO)}, pages 2035--2040. IEEE, 2015.

\bibitem{Rhinehart2016}
Nicholas Rhinehart and Kris~M. Kitani.
\newblock {Learning Action Maps of Large Environments via First-Person Vision}.
\newblock In {\em 2016 IEEE Conference on Computer Vision and Pattern
  Recognition (CVPR)}, pages 580--588. IEEE, jun 2016.

\bibitem{Grabner2011}
Helmut Grabner, Juergen Gall, and Luc {Van Gool}.
\newblock {What makes a chair a chair?}
\newblock In {\em 2011 IEEE Conference on Computer Vision and Pattern
  Recognition (CVPR)}, pages 1529--1536. IEEE, jun 2011.

\bibitem{zhu2016inferring}
Yixin Zhu, Chenfanfu Jiang, Yibiao Zhao, Demetri Terzopoulos, and Song-Chun
  Zhu.
\newblock {Inferring forces and learning human utilities from videos}.
\newblock In {\em Proceedings of the IEEE Conference on Computer Vision and
  Pattern Recognition}, pages 3823--3833, 2016.

\bibitem{wu2020chair}
Hongtao Wu, Deven Misra, and Gregory~S Chirikjian.
\newblock {Is that a chair? imagining affordances using simulations of an
  articulated human body}.
\newblock In {\em 2020 IEEE International Conference on Robotics and Automation
  (ICRA)}, pages 7240--7246. IEEE, 2020.

\bibitem{wang2017binge}
Xiaolong Wang, Rohit Girdhar, and Abhinav Gupta.
\newblock {Binge watching: Scaling affordance learning from sitcoms}.
\newblock In {\em Proceedings of the IEEE Conference on Computer Vision and
  Pattern Recognition}, pages 2596--2605, 2017.

\bibitem{li2019putting}
Xueting Li, Sifei Liu, Kihwan Kim, Xiaolong Wang, Ming-Hsuan Yang, and Jan
  Kautz.
\newblock {Putting humans in a scene: Learning affordance in 3d indoor
  environments}.
\newblock In {\em Proceedings of the IEEE Conference on Computer Vision and
  Pattern Recognition}, pages 12368--12376, 2019.

\bibitem{Jiang2016}
Yun Jiang, Hema~S Koppula, and Ashutosh Saxena.
\newblock Modeling 3d environments through hidden human context.
\newblock {\em IEEE Transactions on Pattern Analysis and Machine Intelligence},
  38:2040--2053, 2016.

\bibitem{ruiz2020geometric}
Eduardo Ruiz and Walterio Mayol-Cuevas.
\newblock {Geometric affordance perception: leveraging deep 3D saliency with
  the Interaction Tensor}.
\newblock {\em Frontiers in Neurorobotics}, 14:45, 2020.

\bibitem{pavlakos2019expressive}
Georgios Pavlakos, Vasileios Choutas, Nima Ghorbani, Timo Bolkart, Ahmed A~A
  Osman, Dimitrios Tzionas, and Michael~J Black.
\newblock {Expressive body capture: 3d hands, face, and body from a single
  image}.
\newblock In {\em Proceedings of the IEEE/CVF Conference on Computer Vision and
  Pattern Recognition}, pages 10975--10985, 2019.

\bibitem{Zhang_2020_CVPR}
Yan Zhang, Mohamed Hassan, Heiko Neumann, Michael~J Black, and Siyu Tang.
\newblock {Generating 3D People in Scenes Without People}.
\newblock In {\em The IEEE/CVF Conference on Computer Vision and Pattern
  Recognition (CVPR)}, 2020.

\bibitem{hassan2019resolving}
Mohamed Hassan, Vasileios Choutas, Dimitrios Tzionas, and Michael~J Black.
\newblock {Resolving 3D human pose ambiguities with 3D scene constraints}.
\newblock In {\em Proceedings of the IEEE/CVF International Conference on
  Computer Vision}, pages 2282--2292, 2019.

\bibitem{zhang2020place}
Siwei Zhang, Yan Zhang, Qianli Ma, Michael~J Black, and Siyu Tang.
\newblock {PLACE: Proximity learning of articulation and contact in 3D
  environments}.
\newblock In {\em 8th international conference on 3D Vision (3DV
  2020)(virtual)}, 2020.

\bibitem{Zhao2014}
Xi~Zhao, He~Wang, and Taku Komura.
\newblock {Indexing 3D Scenes Using the Interaction Bisector Surface}.
\newblock {\em ACM Transactions on Graphics}, 33(3):1--14, jun 2014.

\bibitem{Peternell2000}
Martin Peternell.
\newblock {Geometric properties of bisector surfaces}.
\newblock {\em Graphical Models}, 62(3):202--236, 2000.

\bibitem{Hu2015}
Ruizhen Hu, Chenyang Zhu, Oliver van Kaick, Ligang Liu, Ariel Shamir, and Hao
  Zhang.
\newblock {Interaction Context (ICON): Towards a Geometric Functionality
  Descriptor}.
\newblock {\em ACM Transactions on Graphics}, 34(4):83:1--83:12, jul 2015.

\bibitem{Hu2016}
Ruizhen Hu, Oliver van Kaick, Bojian Wu, Hui Huang, Ariel Shamir, and Hao
  Zhang.
\newblock {Learning how objects function via co-analysis of interactions}.
\newblock {\em ACM Transactions on Graphics}, 35(4):1--13, jul 2016.

\bibitem{Zhao2016}
Xi~Zhao, Ruizhen Hu, Paul Guerrero, Niloy Mitra, and Taku Komura.
\newblock {Relationship templates for creating scene variations}.
\newblock {\em ACM Transactions on Graphics}, 2016.

\bibitem{Zhao2017}
X.~Zhao, M.~G. Choi, and T.~Komura.
\newblock {Character-object interaction retrieval using the interaction
  bisector surface}.
\newblock {\em Eurographics Symposium on Geometry Processing}, 36(2):119--129,
  may 2017.

\bibitem{Yuksel2015}
Cem Yuksel.
\newblock {Sample Elimination for Generating Poisson Disk Sample Sets}.
\newblock In {\em Computer Graphics Forum}, volume~34, pages 25--32, 2015.

\bibitem{Matterport3D}
Angel Chang, Angela Dai, Thomas Funkhouser, Maciej Halber, Matthias Niessner,
  Manolis Savva, Shuran Song, Andy Zeng, and Yinda Zhang.
\newblock {Matterport3D: Learning from RGB-D Data in Indoor Environments}.
\newblock {\em International Conference on 3D Vision (3DV)}, 2017.

\bibitem{replica19arxiv}
Julian Straub, Thomas Whelan, Lingni Ma, Yufan Chen, Erik Wijmans, Simon Green,
  Jakob~J Engel, Raul Mur-Artal, Carl Ren, Shobhit Verma, Anton Clarkson,
  Mingfei Yan, Brian Budge, Yajie Yan, Xiaqing Pan, June Yon, Yuyang Zou,
  Kimberly Leon, Nigel Carter, Jesus Briales, Tyler Gillingham, Elias Mueggler,
  Luis Pesqueira, Manolis Savva, Dhruv Batra, Hauke~M Strasdat, Renzo~De Nardi,
  Michael Goesele, Steven Lovegrove, and Richard Newcombe.
\newblock {The {R}eplica Dataset: A Digital Replica of Indoor Spaces}.
\newblock {\em arXiv preprint arXiv:1906.05797}, 2019.

\bibitem{hassan2021populating}
Mohamed Hassan, Partha Ghosh, Joachim Tesch, Dimitrios Tzionas, and Michael~J
  Black.
\newblock {Populating 3D scenes by learning human-scene interaction}.
\newblock In {\em Proceedings of the IEEE/CVF Conference on Computer Vision and
  Pattern Recognition}, pages 14708--14718, 2021.

\bibitem{shapiro1965analysis}
Samuel~Sanford Shapiro and Martin~B Wilk.
\newblock {An analysis of variance test for normality (complete samples)}.
\newblock {\em Biometrika}, 52(3/4):591--611, 1965.

\bibitem{franke2012chi}
Todd~Michael Franke, Timothy Ho, and Christina~A Christie.
\newblock {The chi-square test: Often used and more often misinterpreted}.
\newblock {\em American Journal of Evaluation}, 33(3):448--458, 2012.

\bibitem{cramer1946mathematical}
Harald Cramer.
\newblock {The two-dimensional case}.
\newblock In {\em Mathematical Methods of Statistics}, pages 260--290.
  Princeton university press, 1946.

\bibitem{agresti2018introduction}
Alan Agresti.
\newblock {\em {An introduction to categorical data analysis}}, pages 39--41.
\newblock John Wiley \& Sons, 2018.

\bibitem{beasley1995multiple}
T~Mark Beasley and Randall~E Schumacker.
\newblock {Multiple regression approach to analyzing contingency tables: Post
  hoc and planned comparison procedures}.
\newblock {\em The Journal of Experimental Education}, 64(1):79--93, 1995.

\bibitem{vsidak1967rectangular}
Zbyn{\v{e}}k {\v{S}}id{\'{a}}k.
\newblock {Rectangular confidence regions for the means of multivariate normal
  distributions}.
\newblock {\em Journal of the American Statistical Association},
  62(318):626--633, 1967.

\end{thebibliography}

\appendix
\clearpage
\setcounter{figure}{0}  
\setcounter{table}{0}  
\renewcommand\thefigure{S.\arabic{figure}}  
\renewcommand\thetable{S.\arabic{table}}

\begin{center}
{\Large \textbf{AROS: Affordance Recognition with One-Shot Human Stances}}

{\large  Supplementary material }
\end{center}

\section{Training Interactions}\label{append:trained_interactions}
\begin{figure}[h]
    \centering
    \includegraphics[width=.7\linewidth]{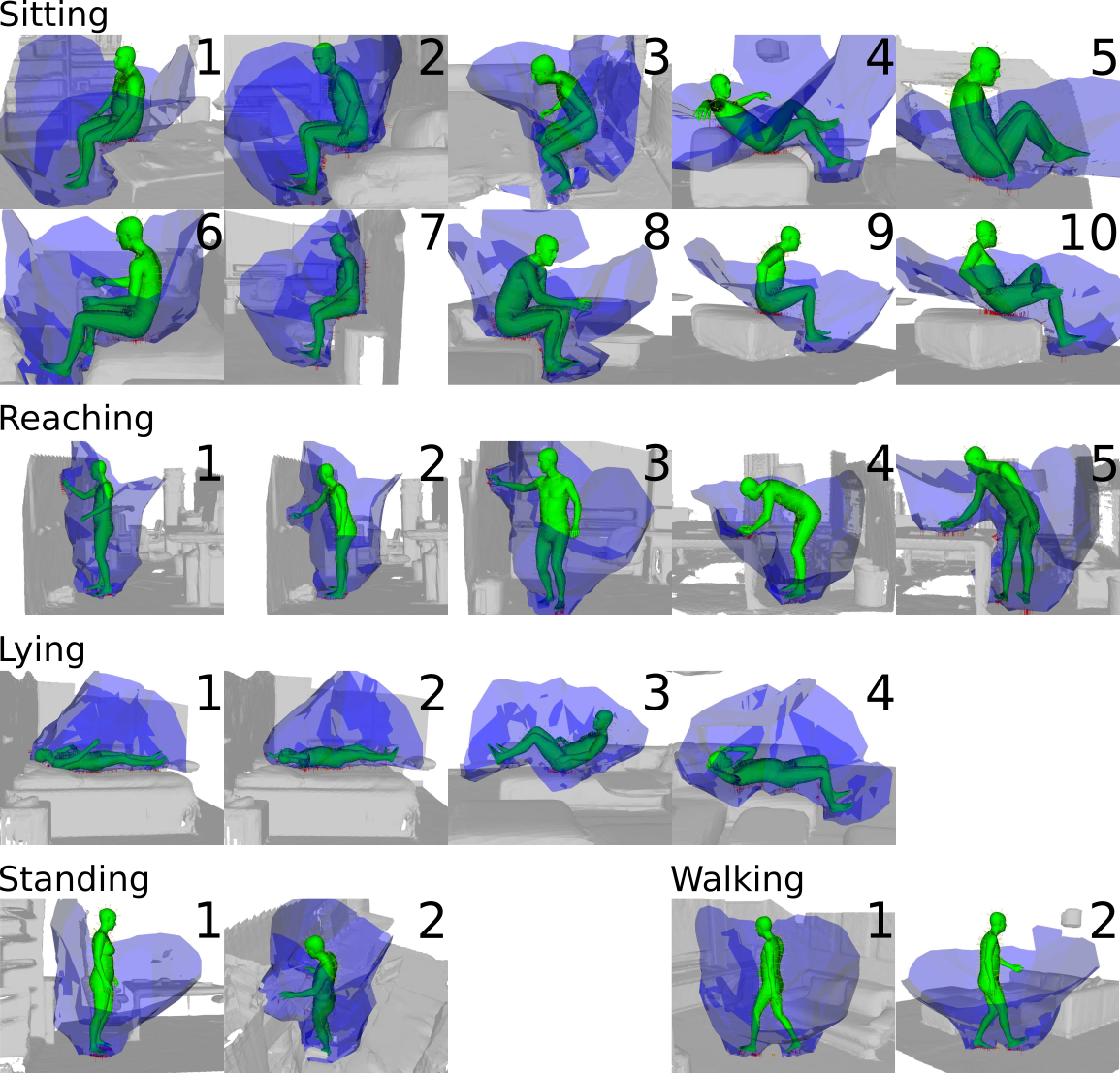}
    \caption{Interactions trained. 5 different categories of human-environment configurations. These are the 23 different interactions used for training in our experiments }
    \label{fig:interactions_trained}
\end{figure}

We train 5 types of interactions: standing, walking, lying, reaching, and sitting (see \cref{fig:interactions_trained}). Training interactions were obtained from recordings in the PROX dataset \cite{hassan2019resolving}. The data of each frame of the recordings include a 3D reconstruction of the observed scene, as well as an SMPLX representation \cite{pavlakos2019expressive} of the human interacting within the scene. For each pair ($M_h$, $M_e$), we also get its reference point $p_{train}$ to calculate the AROS descriptor $(\mathcal{V}_{train},\mathcal{C}_{train}, \hat{n}_{train} )$. Training is thus based on the small selection of human-scene pairs per interaction. Note that semantic labels such as "sitting" are ambiguous, as there are various human poses that are considered sitting.  But note that all of these are distinctive, e.g. sitting with trunk upright or sitting with trunk backwards and arms extended. In AROS, we use only one example to train each of these configurations.

\newpage
\clearpage
\section{Spherical Fillers}
As described in the main text, we preprocess the noisy depth scans that are populated with missing scan regions (holes). We deal with this via the appending of spherical fillers.
Spherical fillers with radius $r$ are generated as follows: (1) Sample points from the environment mesh so that they are approximately evenly separated with a distance $6r/9$, (2) calculate the normal vectors of all sampled points, (3) keep samples from where a ray of size $2r$ can be propagated in the opposite direction to the normal without intersecting the environment mesh, (4) generate spheres tangent to the filtered sampled points, with centers in the rays directions, and (5) crop the spherical fillers with the environment mesh to avoid the creation of artifacts on the scanned scene.

\begin{figure}[h]
    \centering
    \includegraphics[width=.85\linewidth]{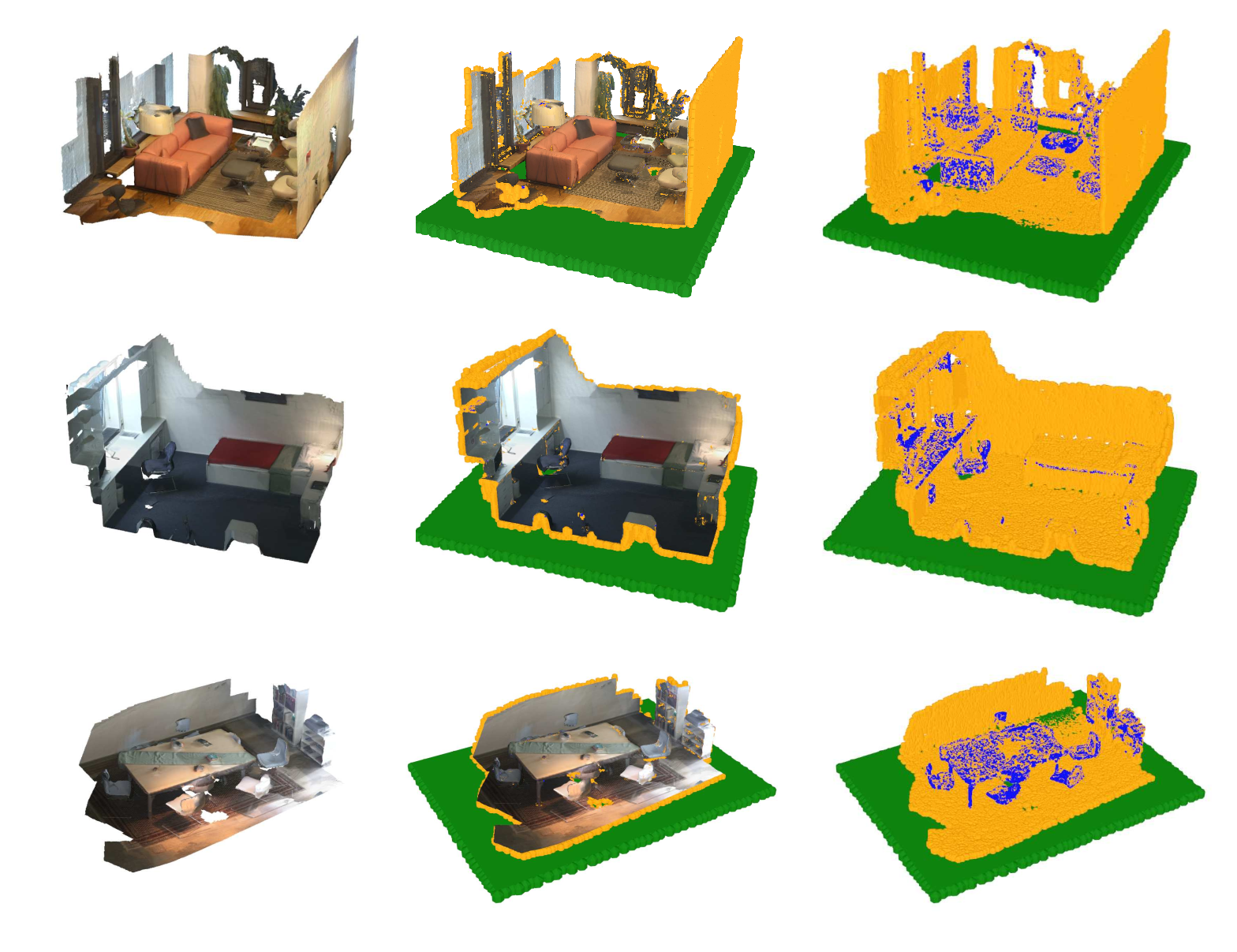}
    \caption{Examples of scene representations enhanced for better collision detection. From left to right: the original mesh definition of a scene, its enhanced version with sphere fillers, and the spherical fillers calculated. Large sphere fillers in green and yellow. Fine sphere fillers in blue}
    \label{fig:more_fullfilling_envs}
\end{figure}

The size and distribution of spherical fillers are dependent on the size and quality of the mesh. We empirically generate three sets of spherical fillers in all scenes from all three datasets as follows: The first, a set of spherical fillers of $r=7[cm]$, aims to occupy the free space generated by large enough elements such as walls, floor, and sofas. The second set, spherical fillers with radius $r=3[cm]$, is meant to fill small gaps of free space in the structure of the environment. The last set of spherical fillers with a radius of 12 [cm], generated on the extended bottom face (40 [cm]) of the scene bounding box, was designed to avoid misinterpretation of the boundaries of the scanned scenes. These values were used for all the dataset and provided here for repeatability. Some results are presented in \cref{fig:more_fullfilling_envs}.

\newpage
\clearpage
\section{User Studies}
\begin{figure}[h]
    \centering
    \includegraphics[width=\linewidth]{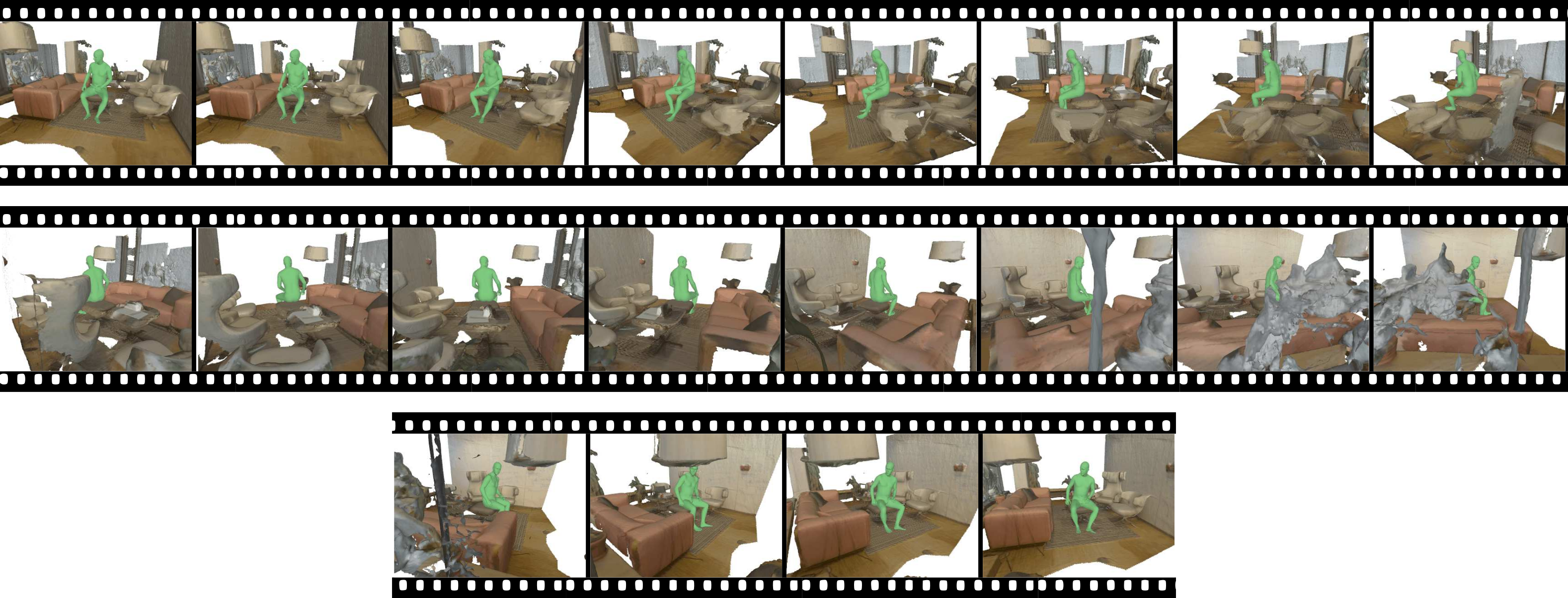}
    \caption{We use animated GIF images to visualize human-scene interactions for the MTurk studies}\label{fig:study_prespectives}
\end{figure}

We designed and conducted two different protocols to evaluate the naturalness of generated human-environment pairs via Amazon Mechanical Turk. We used animated GIF images to present interactions to MTurkers. They observed them from different perspectives with a camera positioned to a moderated distance with respect the human body as shown in \cref{fig:study_prespectives}. In both protocols, every Mturker perform 11 evaluations per task with 2 control questions to remove unreliable assessors performing random answering. The control questions were selected to be as obvious as possible to be easy to decide for someone following the instructions. No MTurker was permitted to participate more than once.

\begin{figure}[h]
    \centering
     \includegraphics[width=.80\linewidth]{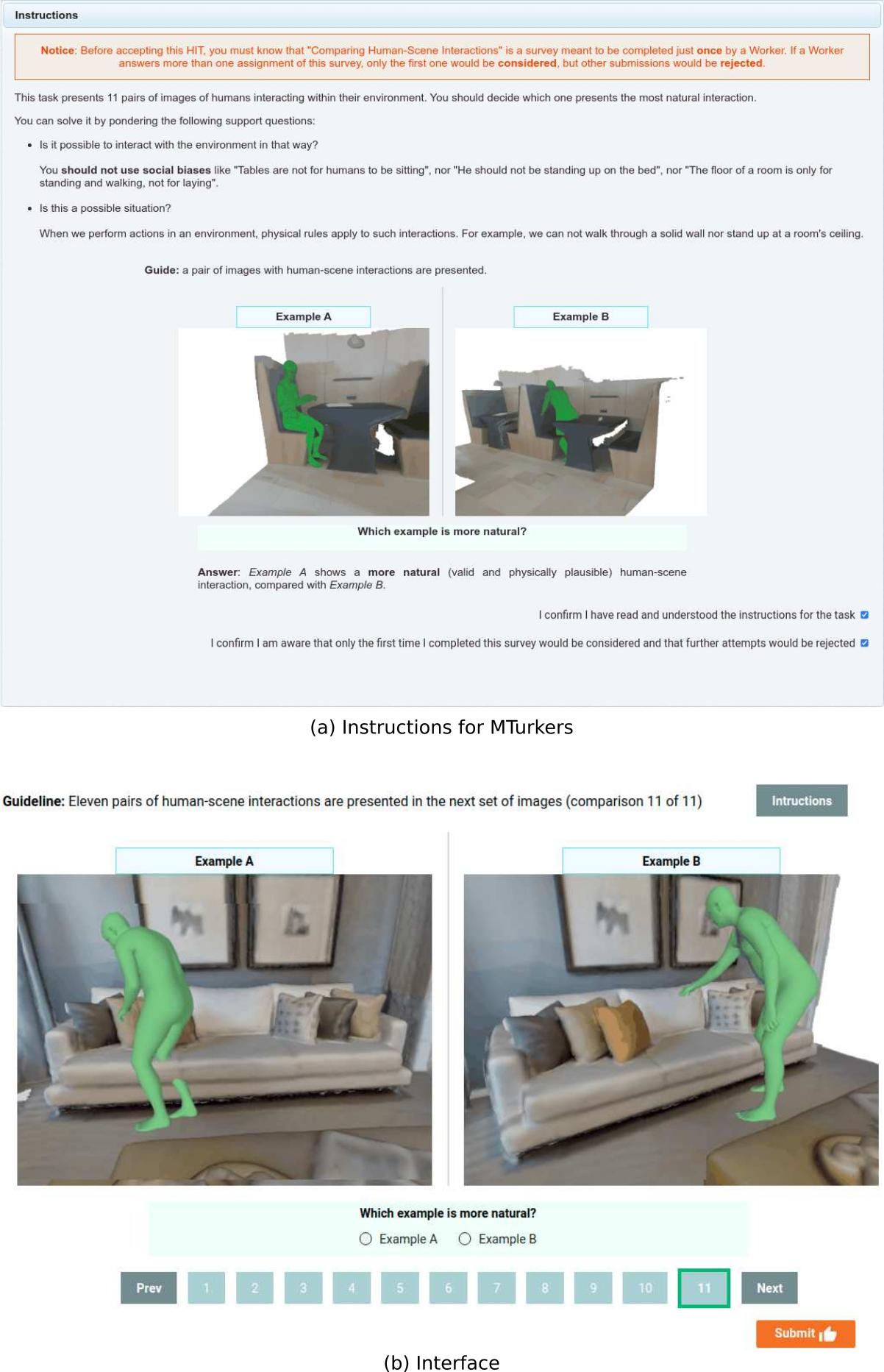}
    \caption{Side by Side Evaluation Study}\label{fig:study_side_by_side}
\end{figure}

For the side by side evaluation study, body-environment pairs generated by both compared approaches \cite{zhang2020place} were presented to MTurkers for a direct comparison. To encourage a fair test, we ask both approaches to generate interaction around same scene locations with a tolerance of 1[m] around them. Left and right placement on the interface was randomized to prevent location bias. Our evaluation instrument, which includes instructions and the interface for MTurkers, is presented in \cref{fig:study_side_by_side}.

\begin{figure}[h]
    \centering
    \includegraphics[width=.8\linewidth]{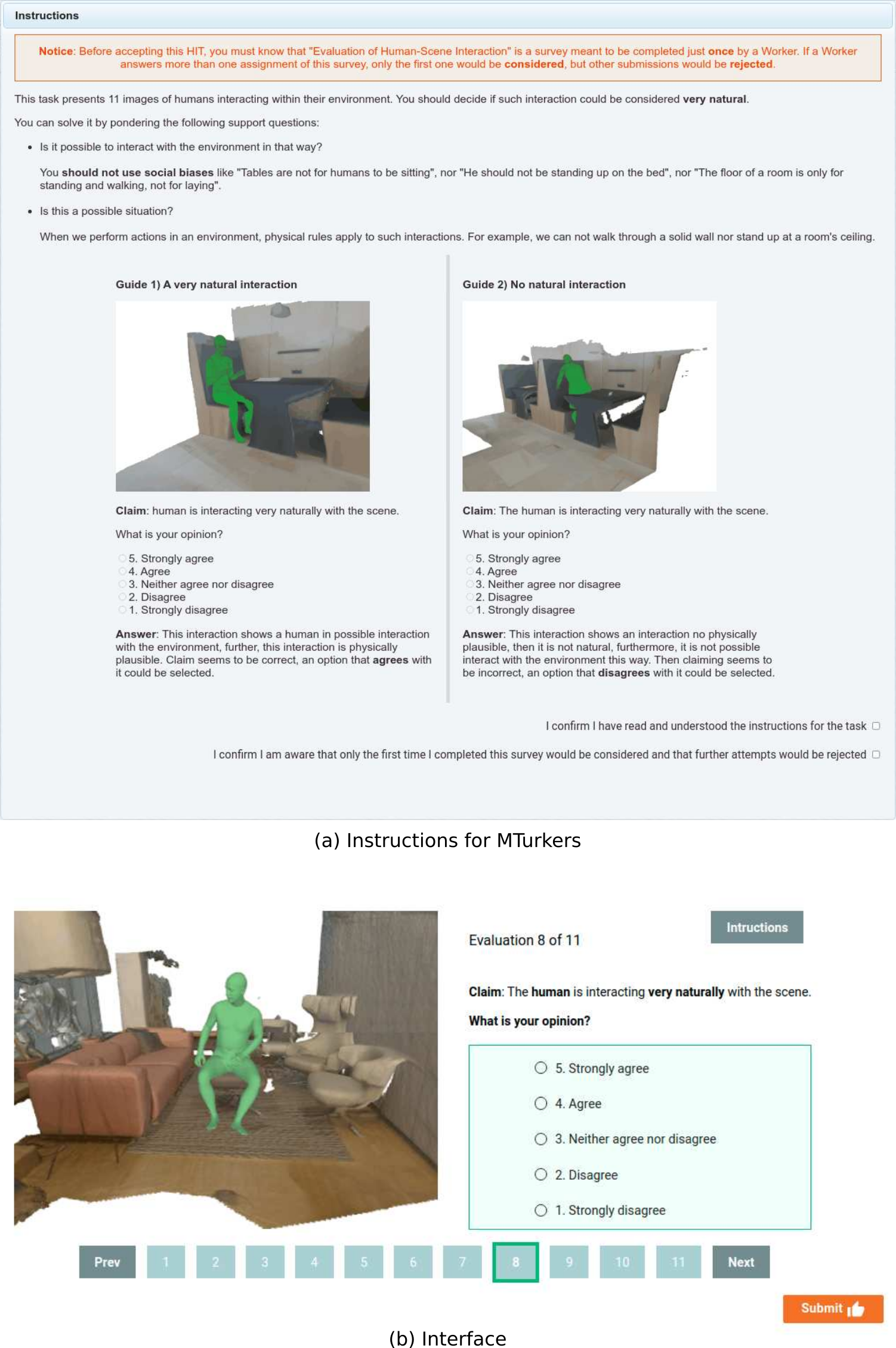}
    \caption{Individual Evaluation Study}\label{fig:study_individual}
    
\end{figure}

For our individual evaluation study, only one interaction at a time per compared method was presented to an MTuker. The set of instructions given and the interface used to evaluate the interactions through a Likert scale is presented in \cref{fig:study_individual}.

\newpage
\clearpage

\section{More on Qualitative Results}
\begin{figure}[h]
    \centering
    \includegraphics[width=.8\linewidth]{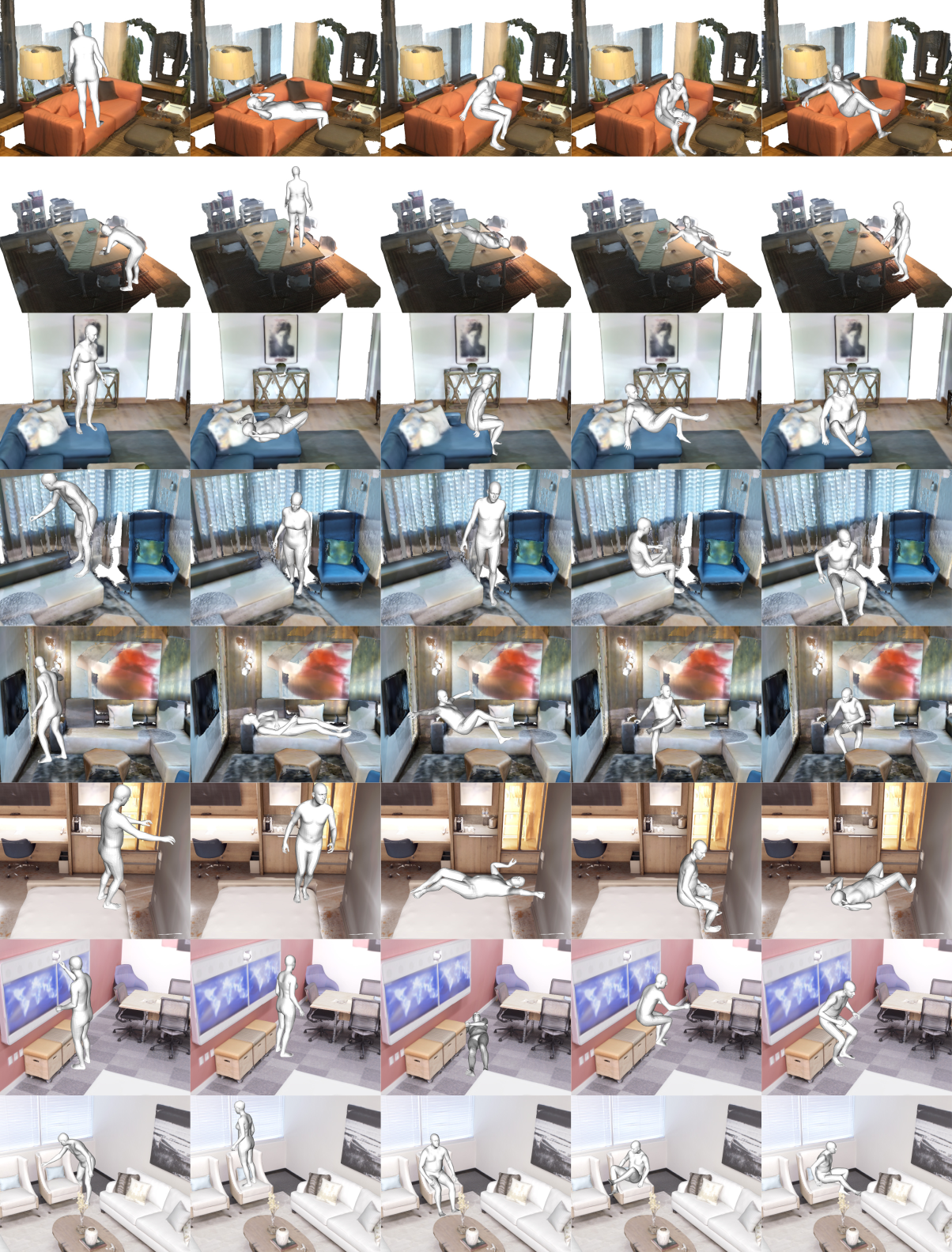}
    \caption{Further example human interactions generated by AROS. Several affordances are detected in the same location}\label{fig:s_multiple_detections}
\end{figure}

AROS answers the question \textit{what can a human perform here?} by detecting different affordances around the same location in an environment. Examples of this capability are presented in \cref{fig:s_multiple_detections}. More positive detections happened, but we presented only 4 in the figure for ease of legibility.

\begin{figure}[h]
    \centering
    \includegraphics[width=.8\linewidth]{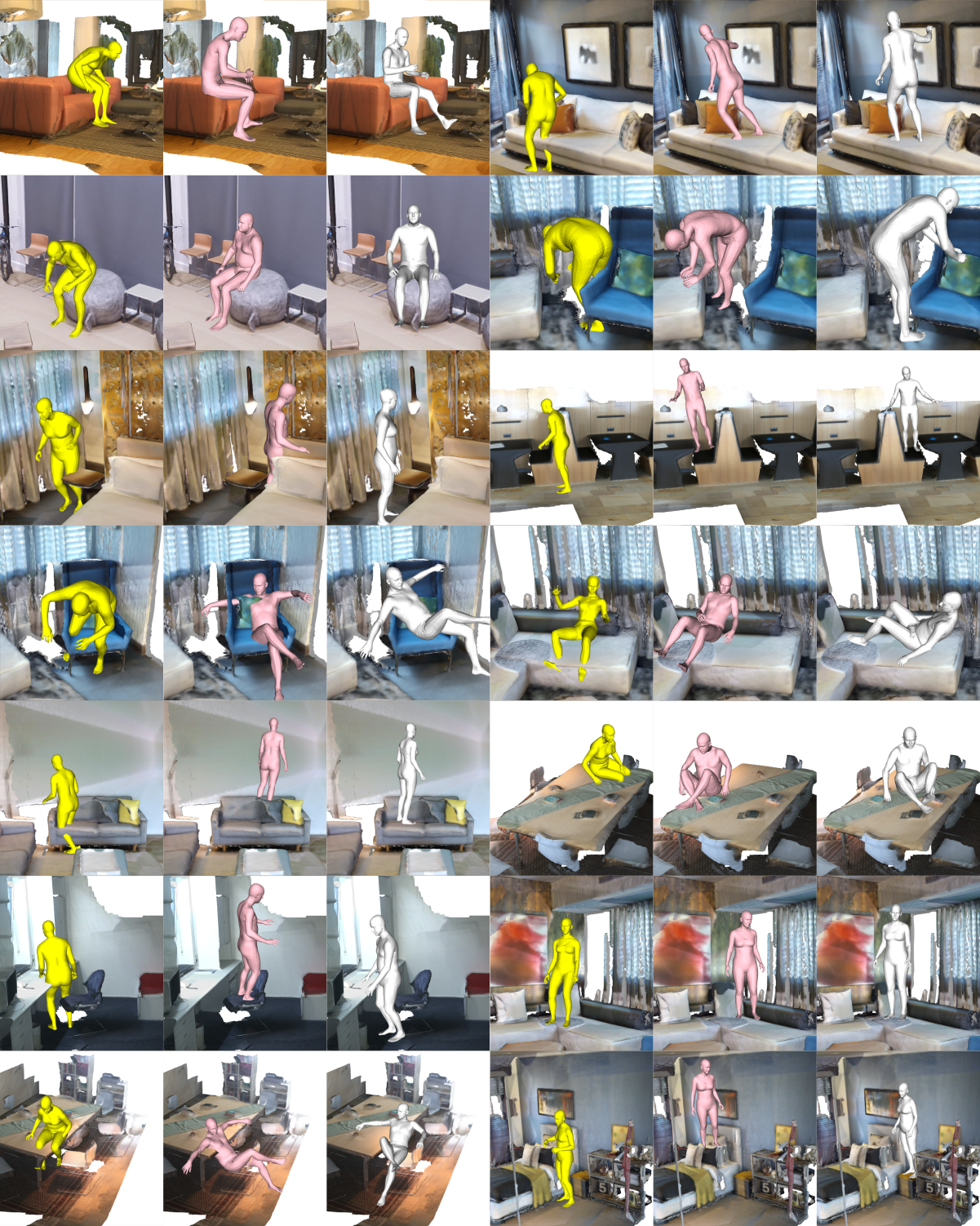}
    \caption{Human poses generated by PLACE (yellow), POSA (pink), and AROS (silver) on challenging locations in the scenes. AROS can often generate multiple and more plausible instances of the same affordance per location volume, here we show two examples }\label{fig:s_comparison_place}
\end{figure}

\cref{fig:s_comparison_place} shows qualitative comparisons between PLACE (\cite{zhang2020place}), POSA (\cite{hassan2021populating}), and our results. It is not possible to control the type of interaction we want PLACE to generate.  In this set of examples, we asked PLACE to generate a human-environment configuration. Then, we classified the resulting interaction into one of the 5 affordances and requested POSA and AROS to detect and generate the same type of interaction.


\newpage
\clearpage

\section{Failure Cases}

Our evaluations show that bodies generated by AROS with AdvOptim optimization are natural and physically plausible. Nonetheless, it can still produce some failure cases with evident penetration in the environment surface.

\begin{figure}[h]
    \centering
    \includegraphics[width=.8\linewidth]{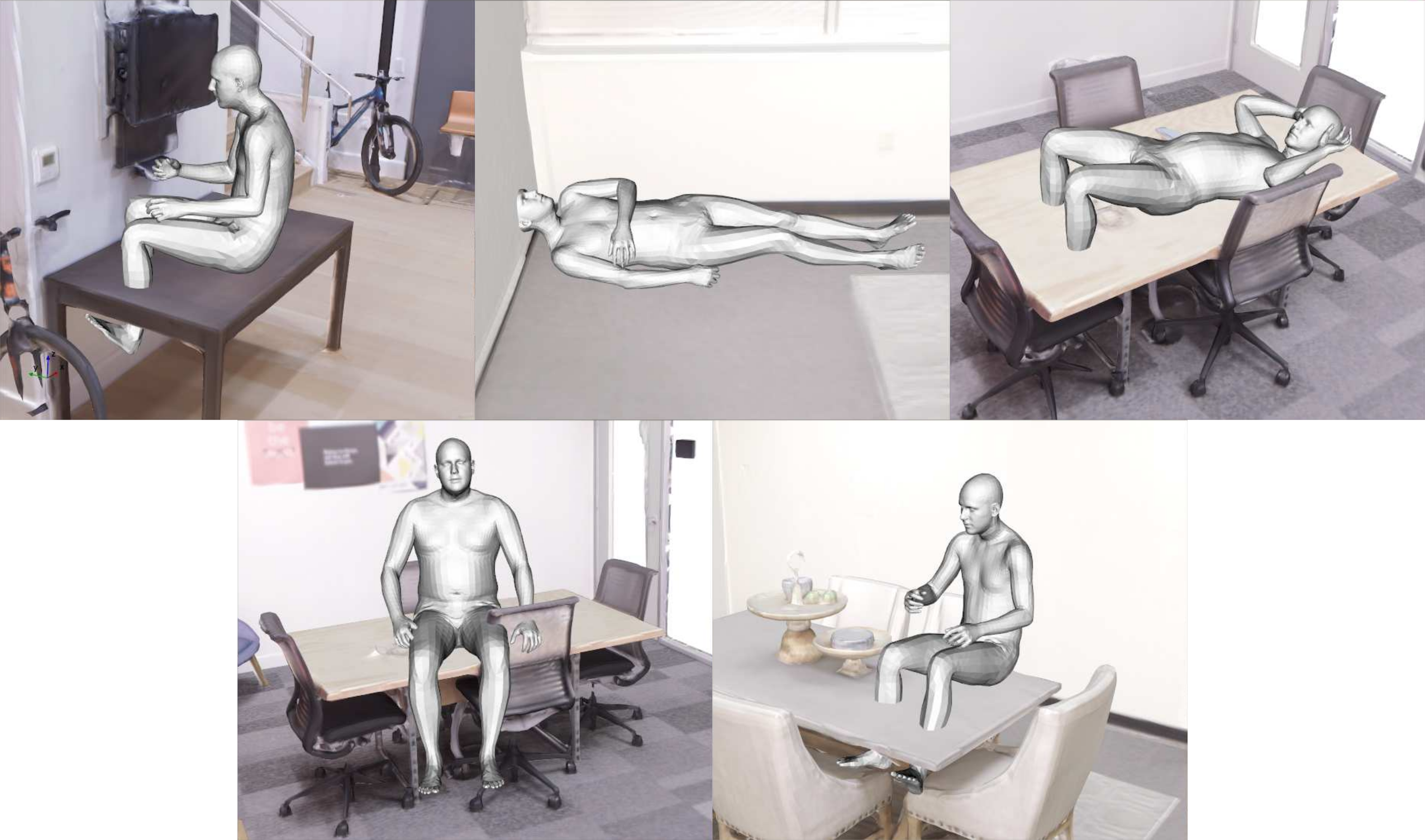}
    \caption{Examples of failure cases}
    \label{fig:failure_cases}
\end{figure}

There are two inter-related reasons for the failure cases. These are the density of \textit{clearance vectors} and the density of the SDF values. Low densities in these parameters lead to being incapable of detecting a noticeable body-environment collision as they fail to properly characterize the space around thin structures in the scene. In our experiments, failures were mainly seen on tables when looking for positions capable of supporting a "sitting" interaction, as shown in \cref{fig:failure_cases}. Increasing densities for \textit{clearance vectors} and/or the SDF will need to be balanced wrt compute time at inference.


\end{document}